\documentclass{article}

\usepackage{microtype}
\usepackage{graphicx}
\usepackage{subfigure}
\usepackage{booktabs} 

\usepackage[utf8]{inputenc}
\usepackage{natbib}
\usepackage{xcolor}
\usepackage{enumitem}

\usepackage{amsmath}
\usepackage{amssymb}
\usepackage{mathtools}
\usepackage{amsthm}
\usepackage{hyperref}
\usepackage{soul}

\usepackage[accepted]{icml2024}

\usepackage{amsmath,amsfonts,bm}

\def\eqref#1{equation~\ref{#1}}

\def\1{\bm{1}}

\def\rvw{{\mathbf{w}}}

\def\rmC{{\mathbf{C}}}

\def\rmE{{\mathbf{E}}}

\def\rmG{{\mathbf{G}}}

\def\rmI{{\mathbf{I}}}

\def\rmP{{\mathbf{P}}}

\def\rmV{{\mathbf{V}}}
\def\rmW{{\mathbf{W}}}
\def\rmX{{\mathbf{X}}}

\def\ermE{{\textnormal{E}}}

\def\ermM{{\textnormal{M}}}

\def\ermO{{\textnormal{O}}}

\def\va{{\bm{a}}}
\def\vb{{\bm{b}}}

\def\mM{{\bm{M}}}

\def\mU{{\bm{U}}}
\def\mV{{\bm{V}}}

\def\mX{{\bm{X}}}
\def\mY{{\bm{Y}}}

\DeclareMathAlphabet{\mathsfit}{\encodingdefault}{\sfdefault}{m}{sl}
\SetMathAlphabet{\mathsfit}{bold}{\encodingdefault}{\sfdefault}{bx}{n}

\def\gD{{\mathcal{D}}}

\def\gL{{\mathcal{L}}}

\def\gP{{\mathcal{P}}}

\def\gX{{\mathcal{X}}}
\def\gY{{\mathcal{Y}}}

\def\sU{{\mathbb{U}}}

\newcommand{\Pa}{\mathrm{PA}_} 

\DeclareMathOperator*{\argmin}{arg\,min}

\theoremstyle{plain}
\newtheorem{theorem}{Theorem}[section]

\newtheorem{lemma}[theorem]{Lemma}

\theoremstyle{definition}
\newtheorem{definition}[theorem]{Definition}

\theoremstyle{remark}

\icmltitlerunning{Optimal Transport for Structure Learning Under Missing Data}

\begin{document}
\twocolumn[
\icmltitle{Optimal Transport for Structure Learning Under Missing Data}

\icmlsetsymbol{equal}{*}

\begin{icmlauthorlist}
\icmlauthor{Vy Vo}{yyy,ccc}
\icmlauthor{He Zhao}{ccc}
\icmlauthor{Trung Le}{yyy}
\icmlauthor{Edwin V. Bonilla}{ccc}
\icmlauthor{Dinh Phung}{yyy,vvv}
\end{icmlauthorlist}

\icmlaffiliation{yyy}{Monash University, Australia}
\icmlaffiliation{ccc}{CSIRO's Data61, Australia}
\icmlaffiliation{vvv}{VinAI Research, Vietnam}

\icmlcorrespondingauthor{Vy Vo}{v.vo@monash.edu}

\icmlkeywords{Machine Learning, ICML}
\vskip 0.3in
]

\printAffiliationsAndNotice{} 
\begin{abstract}
Causal discovery in the presence of missing data introduces a chicken-and-egg dilemma. While the goal is to recover the true causal structure, robust imputation requires considering the dependencies or, preferably, causal relations among variables. Merely filling in missing values with existing imputation methods and subsequently applying structure learning on the complete data is empirically shown to be sub-optimal. To address this problem, we propose a score-based algorithm for learning causal structures from missing data based on optimal transport. This optimal transport viewpoint diverges from existing score-based approaches that are dominantly based on expectation maximization. We formulate structure learning as a density fitting problem, where the goal is to find the causal model that induces a distribution of minimum Wasserstein distance with the observed data distribution. Our framework is shown to recover the true causal graphs more effectively than competing methods in most simulations and real-data settings. Empirical evidence also shows the superior scalability of our approach, along with the flexibility to incorporate any off-the-shelf causal discovery methods for complete data. 

\end{abstract}
\section{Introduction}\label{sect:intro}
Discovering  causal relationships among different variables holds great significance in many scientific disciplines \cite{sachs2005causal,richens2020improving,wang2020causal,zhang2013integrated}. The gold standard approach to this problem is through randomized controlled experiments, which are however often infeasible due to various ethical and practical constraints. There have thus been ongoing efforts towards causal discovery from observational data \citep{spirtes2000causation,chickering2002optimal,heckerman2006bayesian,zheng2018dags,yu2019dag,glymour2019review,zheng2020learning,zhao2024bayesian,bonilla2024variational}. Whereas existing structure learning algorithms mostly deal with complete data, practical real-world data are often messy, potentially with multiple missing values. Eliminating samples affected by `missingness' and solely analyzing the observed cases is undesirable, since not only would it decrease the sample size but also introduce bias to the estimations \citep{tu2019causal,mohan2021graphical}. 
Another strategy is to impute the missing values by, e.g., modeling the joint data distribution. The key challenge of imputation lies in effectively modeling this distribution when dealing with a substantial number of missing values, 
which may arise form different missing-data mechanisms \citep{little2019statistical}. More importantly, as proposed by 
\citet{mohan2021graphical}, 
the missing process can also be modeled with causal graphs and, 
without considering the underlying causes of missingness, learning an imputation model from observed data is prone to bias 
\citep{kyono2021miracle}. 
Thus, dealing with missing values benefits substantially from having a causal graph, which is unknown in practice and needed to be learned from data, potentially in the presence of missing values. Evidently, this poses a chicken-and-egg problem.

\begin{figure}[t!]
    \centering
    \includegraphics[width=\linewidth]{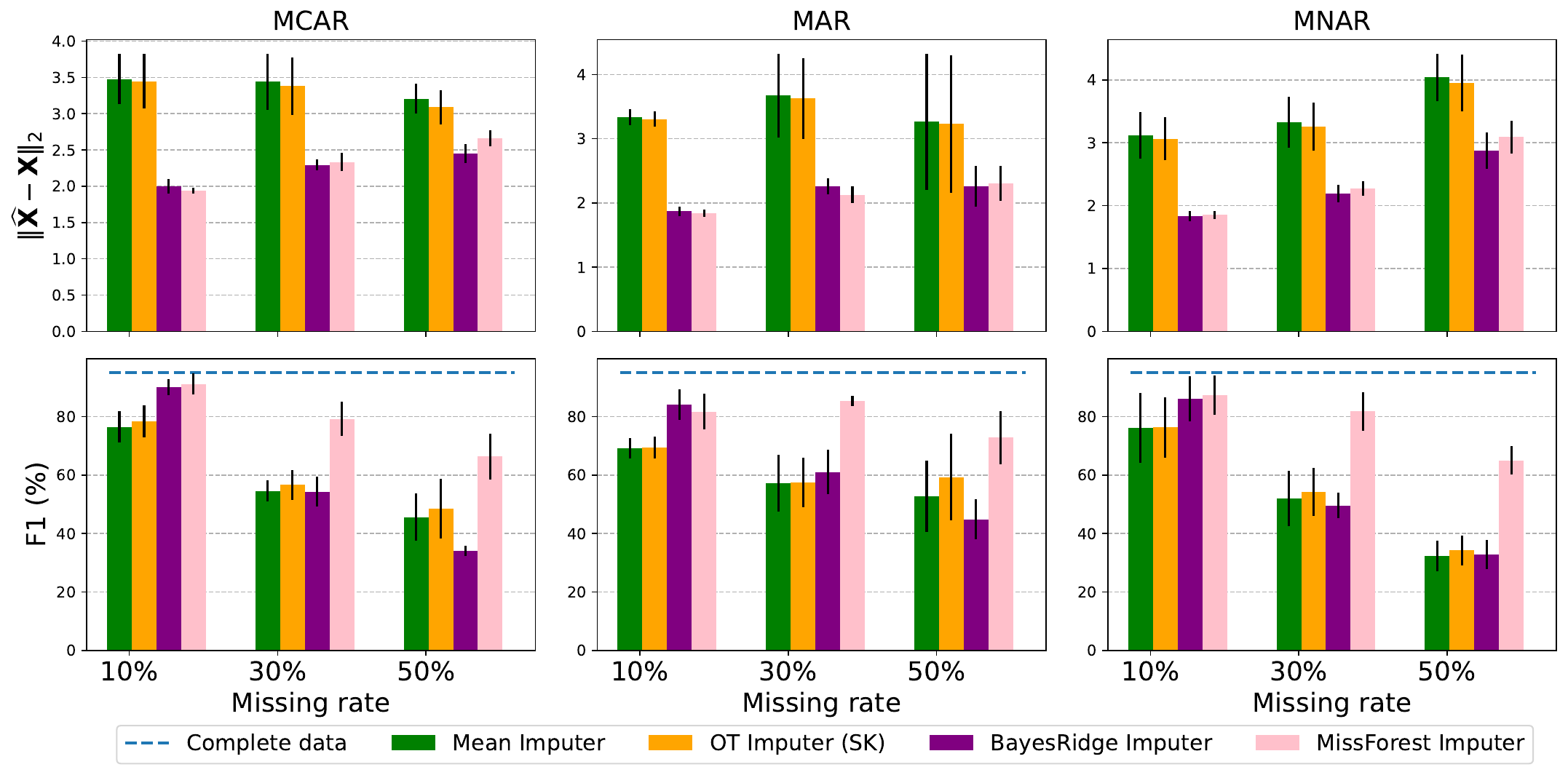}
    \vspace{-5mm}
    \caption{Visualization of the quality of imputation vs. causal discovery performance. Better imputation in terms of reconstruction quality does not always imply more accurate structure learning.}
    \label{fig:intro}
\end{figure}

A straightforward approach to dealing with our problem is to impute the missing values and subsequently apply any existing causal discovery method. This is however not an effective strategy. Figure \ref{fig:intro} illustrates the quality of imputation (by Euclidean distance) in relation to causal discovery performance (by F1 score) across different missing rates and mechanisms. We study $3$ popular imputation methods: Mean imputation, Optimal Transport (OT) imputation (with Sinkhorn divergence) \citep{muzellec2020missing}, and Multivariate imputation with Bayes Ridge regression and Random Forest regression (MissForest) \citep{van2000multivariate,pedregosa2011scikit}. It is seen that better imputation does 
not necessarily lead to better causal discovery, as shown more obviously at higher missing rates. In other words, ``good" imputation in terms of reconstruction does not guarantee ``good" imputation for causal discovery. It is possible that the filled-in data distribution encodes a different set of independence constraints, resulting in a distribution compatible with a  causal graph that is different from the true one. The sub-optimality of such a naive approach has also been reported in  prior work \citep{kyono2021miracle,gao2022missdag}. This motivates the development of a joint end-to-end machinery dedicated to structure learning under missing data.

Depending on the base causal discovery algorithm (designed for complete data), existing methods can also be categorized as either constraint-based or score-based. Constraint-based methods are mainly built up on the PC algorithm, which exploits (conditional) independence tests \citep{strobl2018fast,tu2019causal,gain2018structure}. Score-based causal discovery has recently taken off, owning to advances in exact characterization of acyclicity of the graph \citep{zheng2018dags,yu2019dag,bello2022dagma}. Extension of score-based approach in the context of missing data is nonetheless less studied\footnote{A review of other related works is provided in Appendix \ref{sect:rwork}.}. A notable method in this line of research is MissDAG \citep{gao2022missdag}, which proposes an EM-style algorithm \citep{Dempster1977} to estimate the graph. In the E step, MissDAG imputes the missing entries by modeling a posterior distribution over the missing part of the data. This gives rise to the expected log-likelihood of the complete data, which MissDAG maximizes to estimate the model parameters in the M step. MissDAG has shown a significantly improved performance over constraint-based methods. However, the framework suffers from two key drawbacks. First, the time inefficiency issue of vanilla EM hinders the scalability of MissDAG. Second, a large part of the work focuses on linear Gaussian models where the posterior exists in closed forms. In the non-Gaussian and non-linear cases, MissDAG resorts to rejection sampling, which introduces extra computational overhead. Furthermore, approximate inference may compromise the estimation accuracy greatly, especially in real-world settings where the model tends to be misspecified. 

\paragraph{Contributions.}  In this work, we introduce \textbf{OTM} - an \textbf{O}ptimal \textbf{T}ransport framework for learning causal graphs under \textbf{M}issing data.  From the viewpoint of optimal transport \citep[OT,][]{villani2009optimal}, we fit structure learning into a general landscape of density fitting problems, to which OT has proved successful \citep{arjovsky2017wasserstein,tolstikhin2017wasserstein,vuong2023vector,vo2024parameter}. The high-level idea is to minimize the Wasserstein distance between the model distribution (often in a parametric family) and the empirical data distribution. There are two key properties that makes OT an attractive solution to our problem: (1) the Wasserstein distance is a metric, thus providing a more geometrically meaningful distance between two distributions than the standard $f-$divergences; (2) the Wasserstein estimator is empirically shown in past studies to be more robust to misspecifications \citep{bernton2019parameter,vo2024parameter} and several practical settings where the maximum likelihood estimate fails, e.g., when the model is a singular distribution or the two distributions have little or no overlapping support \citep{bassetti2006minimum,canas2012learning,peyre2017computational,arjovsky2017wasserstein}. 

Our proposal OTM is a flexible framework in the sense that (1) OTM can accommodate any existing score-based causal discovery algorithm for complete data, making it able to handle missing values, and (2) we make no assumptions about the model forms nor the missing mechanism. In Section \ref{sect:method}, we present our theoretical development that renders a tractable formulation of the OT cost to optimize the model distribution on the missing data. In the remainder of the paper, we provide a comprehensive study of the non-linear scenarios on both synthetic and real datasets. Most of the experiments demonstrate the best/second-best performances of OTM on accurately recovering the true causal structures, along with superior scalability when the graph complexity increases.   

\section{Preliminaries}
We first introduce the notations and basic concepts in the paper. We reserve bold capital letters (e.g., $\rmG$) for notations related to graphs. We use calligraphic letters (e.g., $\gX$) for spaces and bold upper case (e.g., $\mX$) for matrices. We also use upper case letters (e.g., $X$) for random variables and lower case letters (e.g., $x$) for values.  Finally, we use $[d]$ to denote a set of integers $\{1, 2, \cdots, d\}$.

\subsection{Structural Causal Model}
A directed graph $\rmG = (\rmV, \rmE)$ consists of a set of nodes $\rmV$ and an edge set $\ermE \subseteq \rmV^2$  of ordered pairs of
nodes with $(v, v) \notin \rmE$ for any $v \in \rmV$ (one without self-loops). For a pair of nodes $i,j$ with $(i,j) \in \ermE$, there is an arrow pointing from $i$ to $j$ and we write $i \rightarrow j$. Two nodes $i$ and $j$ are adjacent if either $(i,j) \in \rmE$ or $(j,i) \in \rmE$. If there is an arrow from $i$ to $j$ then $i$ is a parent of $j$ and $j$ is a child of $i$. Let $\Pa{X_i}$ denote the set of variables associated with parents of node $i$ in $\rmG$. 

The causal relations among $d$ variables is characterized via a \textit{structural causal model} \citep[SCM,][]{pearl2009causality} over the tuple $\langle U, X, f \rangle$ that generally consists of a sets of assignments 
\begin{align}\label{eq:scm}
    X_i := f_i \big(\Pa{X_i}, U_i \big), \quad i \in \rmV, 
\end{align}
where $U_i$ is an exogenous variable assumed to be mutually independent with variables $\{U_1, \cdots, U_d\} \backslash U_i$. Given a joint distribution over the exogenous variables $P(U_1, \cdots, U_d)$, the functions $f = \left[f_i\right]_{i=1}^{d}$ define a joint distribution $P(X)$ over the endogenous variables $X$. Each SCM induces a causal graph $\rmG$, which we assume in this work to be a \textit{directed acyclic graph} (DAG). We also make standard causal discovery assumptions: (1) the distribution $P(X)$ and the induced graph $\rmG$ satisfies the Markov property \cite{pearl2009causality} and (2) there are no latent confounders of the observed variables. In this work, we focus on the additive noise models \citep[ANMs,][]{peters2014causal,hoyer2008nonlinear} where the graph $\rmG$ may be uniquely identified. In ANMs, (\ref{eq:scm}) takes the form $X_i := f_i(\Pa{X_i}) + U_i, \quad i \in \rmV$.

\subsection{DAG Learning and Acyclicity Characterization}

The task of structure learning is to discover the DAG $\rmG$ underlying a given joint distribution $P(X)$. In this work, we follow a score-based approach where one defines a score or loss function $\gL(f; \mX)$ that measures the ``goodness of fit'' of a candidate SCM to a data matrix $\mX$. The goal is to learn a DAG that minimizes $\gL(f; \mX)$, which can be any such as least squares, logistic loss or log-likelihood. The problem is generally NP-hard due to the combinatorial nature of the space of DAGs \citep{chickering1996learning,chickering2004large}. Thanks to the existing proposals of smooth non-convex characterization of acyclicity \citep{zheng2018dags,yu2019dag,bello2022dagma}, score-based structure learning has successfully been cast as a continuous optimization problem that admits the use of gradient-based optimizers. 

Here the graphical structure is represented by a weighted adjacency matrix $\rmW \in \mathbb{R}^{d \times d}$ where $i \rightarrow j$ if and only if $\rmW_{ij} \ne 0$. In the linear case, we can use $\rmW = [\rvw_1], \cdots, [\rvw_d]$ to define a SCM where (\ref{eq:scm}) takes the form $X_i = \rvw^T_i X + U_i$.  In the non-linear case, \citet{zheng2018dags} propose to consider each $f_j$ in the Sobolev space of square-integrable functions, denoted by $H^1(\mathbb{R}^d) \subset L^2(\mathbb{R}^d)$, whose derivatives are also square integrable. Let $\partial_{i}f_j$ denote the partial derivative of $f_j$ w.r.t $X_i$. The dependency structure implied by the SCM $f=[f_i]_{i=1}^d$ can be encoded in a matrix $\rmW = \rmW(f) \in \mathbb{R}^{d \times d}$ with entries $[\rmW(f)]_{ij} := \Vert \partial_{i}f_j \Vert_{2}$, where $\Vert \cdot \Vert_{2}$ is the $L^2$ norm. Such construction is motivated from the property that $f_j$ is independent of $X_i$ if and only if $\Vert \partial_{i}f_j \Vert_{2} = 0$ \citep{rosasco2013nonparametric}.  In practice, $f$ is approximated with a flexible family of parameterized function such as deep neural networks. 

Learning the discrete graph $\rmG$ is now equivalent to optimizing for $\rmW$ in the continuous space of $d \times d$ real matrices. This allows for encoding the acyclicity constraint in $\rmW$ during optimization via a regularizer $R(\rmW)$ such that $R(\rmW) = 0$ if and only $\rmW \in$ DAGs. We briefly review popular acyclicity characterizations in the following. 

\begin{itemize}
\item \textbf{NOTEARS} \cite{zheng2018dags}:
$\quad R(\rmW) = \textrm{trace}\left[\exp (\rmW \circ \rmW)\right] - d.$
\item \textbf{Polynomial} \citep{yu2019dag}: $R(\rmW) = \textrm{trace}\left[(\rmI + \alpha \rmW \circ \rmW)^{d}\right] - d$ for $\alpha > 0$.
\item \textbf{DAGMA} \citep{bello2022dagma}: $R(\rmW) = - \log \det (s\rmI - \rmW \circ \rmW) + d \log s$ for $s > 0$.  
\end{itemize}
\vspace{-1em}
where $\circ$ denotes the Hadamard product and $\rmI$ denotes the identity matrix. 

\subsection{Optimal Transport}
Let $\alpha = \sum^{n}_{j=1} a_j \delta_{x_j}$ be a discrete measure with weights $\va$ and particles $x_1, \cdots, x^n \in \gX$ where $\va \in \Delta^{n}$, a $(n-1)-$dimensional probability simplex. Let $\beta = \sum^{n}_{j=1} b_j \delta_{y_j}$ be another discrete measure defined similarly. The Kantorovich \citep{kantorovich2006problem} formulation of the OT distance between two discrete distributions $\alpha$ and $\beta$ is 
\begin{equation}\label{eq:kanto_disc}
    W_c \left(\alpha, \beta \right) := \inf_{\rmP \in \sU(\va,\vb)} \langle \rmC, \rmP \rangle,
\end{equation}
where $\langle \cdot, \cdot \rangle$ denotes the Frobenius dot-product; $\rmC \in \mathbb{R}^{n \times n}_{+}$ is the cost matrix of the transport; $\rmP \in \mathbb{R}^{n \times n}_{+}$ is the transport matrix/plan; $\sU(\va,\vb) := \left\{ \rmP \in \mathbb{R}^{n \times n}_{+} : \rmP \mathbf{1}_{n} = \va, \rmP \mathbf{1}_{n} = \vb \right\} $ is the transport polytope of $\va$ and $\vb$; $\mathbf{1}_{n}$ is the $n-$dimensional column of vector of ones. For arbitrary measures, Eq. (\ref{eq:kanto_disc}) can be generalized as 
\begin{equation}\label{eq:kanto_cont}
    W_c \left(\alpha; \beta\right) := \underset{\Gamma \sim \gP (X \sim \alpha, Y \sim \beta)}{\mathrm{inf}} \mathbb{E}_{(X, Y) \sim \Gamma} \bigl[ c(X,Y)\bigr],
\end{equation}
where the infimum is taken over the set of all joint distributions $\gP (X \sim \alpha, Y \sim \beta)$ with marginals $\alpha$ and $\beta$ respectively. $c: \gX \times \gY \mapsto \mathbb{R}$ is any measurable cost function. If $c(x,y) = D^p(x,y)$ is a distance for $p \le 1$, $W_p$, the $p$-root of $W_c$, is called the $p$-Wasserstein distance.  Finally, for a measurable map $T : \gX \mapsto \gY$, $T\#\alpha$ denotes the push-forward measure of $\alpha$ that, for any measurable set $B \subset \gY$, satisfies $T\#\alpha(B) = \alpha(T^{-1}(B))$. For discrete measures, the push-forward operation is $T\#\alpha = \sum^{n}_{j=1} a_j \delta_{T(x_j)}$.


\subsection{Data with Missing Values}\label{bg:missing}
We consider a dataset of $n$ data samples with $d$ features stored in matrix $\mX = \left[ \mX^{1}, \cdots, \mX^{n} \right]^T \in \mathbb{R}^{n\times d}$. An indicator matrix $\mM \in \{0,1\}^{n \times d}$ is used to encode the missing data so that $\mM^{j}_{i}=1$ indicates that the $i^{\text{th}}$ feature of sample $j$ is missing and $\mM^{j}_{i}=0$ otherwise. The missing values are initially assigned with Nan (Not a number). We follow Python/Numpy style matrix indexing $\mX[\mM]=\text{Nan}$ to denote the missing data.

The missing mechanism falls into one of the three common categories \citep{little2019statistical}: 
missing completely at random (MCAR) where
the cause of missingness is purely random and independent of the data; missing at random (MAR) where the probability of being missing depends only on observed values; missing not at random (MNAR) where the cause of missingness is unobserved. Dealing with MAR or MNAR is more challenging, as missing values may introduce significant biases into the analysis results \citep{mohan2021graphical,kyono2021miracle,jarrett2022hyperimpute}.

\section{Methodology}\label{sect:method}
\subsection{Problem Setup} 
We consider a causal DAG $\rmG(\rmV, \rmE)$ on $d$ random variables $X$ induced by an underlying SCM over the tuple $\langle U,X,f\rangle$. Let $\theta$ denote the parameters of our model, \textbf{including} the weighted adjacency matrix $\rmW$ parameterizing the causal graph and the parameters of the functions approximating $f = [f_i]_{i=1}^{d}$. We denote these functions as $f_{\theta}$. 

We focus on the finite-sample setting with missing data. Given a dataset $\gD = (\mX, \mM)$, let $\ermO \subseteq [d]$ denote the set of entries/variables in each observation $\rmX^{j}$. Let $\mX^{j}_{\ermO} = \left[\mX^{j}_{i} : \mM^{j}_{i} = 0, i \in [d]\right]$ denote the observed part of $\mX^{j}$. Similarly, we have $\mX_{\ermO} =\left[\mX^{j}_{i} : \mM^{j}_{i} = 0, i \in [d], j \in [n] \right]$ denote the entire observed part of the data.  

Let $\mu_{\gD}(\mX_{\ermO}) = n^{-1} \sum_{j=1}^{n} \delta_{\mX^{j}_{\ermO}}$ denote the (incomplete) empirical distribution over $\mX_{\ermO}$. In the classic density-fitting problem where the data is complete, the task is to find $\theta$ that minimizes some distance/divergence between the model distribution and the empirical distribution over the observed data. This problem setup however does not translate smoothly to our setting of missing data since the definition of a model distribution over $\mX_{\ermO}$, denoted as $\mu_{\theta}(\mX_{\ermO})$, is not straightforward. Now we explain how to address this issue. 

With a slight abuse of notation, we use $\mX$ to denote the true (unknown) data matrix. Given a complete data matrix $\mX$, we define $h_m$ as a mechanism to produce missing data, or equivalently to extract the observed part, i.e., $h_m(\mX) := \left[\mX^{j}_{i} : \mM^{j}_{i} = 0, i \in [d], j \in [n] \right] = \mX_{\ermO}$ and $h_m$ operates row-wisely according to $\mM$. Given a sub-part of data $\mX^{j}_{\ermO}$, we construct a mechanism $h_c$ to complete the data such that $h_c(\mX^{j}_{\ermO}) = \widetilde{\mX}^{j} \in \mathbb{R}^{d}$.  For $h_m$ and $h_c$ defined above, we can define the model distribution over $\mX_{\ermO}$ via its reconstruction from the model, that is $\mu_{\theta}(\mX_{\ermO}) := n^{-1} \sum^{n}_{j=1} \delta_{h_m \left\{ f_{\theta} \left[ h_c(\mX^{j}_{\ermO}) \right] \right\}  }$. 

\subsection{Proposed Method}

We propose to learn the graph $\rmG$ using the OT cost as the score function. The goal is to find $\theta$ that minimizes the OT distance between $\mu_{\gD}(\mX_{\ermO})$ and $\mu_{\theta}(\mX_{\ermO})$: 
\begin{align}\label{eq:initial}
\hat{\theta} & = \underset{\theta : f_{\theta} \in H^1(\mathbb{R}^d)}{\argmin} W_c \left[\mu_{\gD}(\mX_{\ermO}),\mu_{\theta}(\mX_{\ermO}) \right]\\
& \text{subject to } \rmW(f) \in \text{ DAGs,} \nonumber
\end{align}
where the OT distance is given as 
\begin{align*}
    & W_c \left[\mu_{\gD}(\mX_{\ermO}), \mu_{\theta}(\mX_{\ermO}) \right] \\
    & = \min_{\rmP \in \sU(\mathbf{1}_{n},\mathbf{1}_{n}) } \sum_{j,k}  c \left\{ \mX^{j}_{\ermO}, h_m \left[ f_{\theta} \left( h_c(\mX^{k}_{\ermO}) \right) \right]  \right\} \rmP^{j,k}.
\end{align*}
Note that $\rmP \in \mathbb{R}_{+}^{n \times n}$ is a coupling matrix and the minimum is taken over all possible couplings in the Birkhoff polytope $\sU(\mathbf{1}_{n}/n,\mathbf{1}_{n}/n)$. The challenge now is how to define the cost $c(\cdot, \cdot)$ between two particles at locations $(j, k)$, two arbitrary sub-vectors of possibly different dimensions. Notice that if the data matrix is complete i.e., each particle is a complete $d-$dimensional vector, the cost function can be naturally defined on the continuous space. 

\begin{definition}\label{def:1} (Cost function for incomplete samples) The transport cost between a particle $j$ from $\mu_{\gD}(\mX_{\ermO})$ and a particle $k$ from $\mu_{\theta}(\mX_{\ermO})$ is given by
    \begin{align}\label{eq:dist}
    & c \left\{ \mX^{j}_{\ermO}, h_m \left[ f_{\theta} \left( h_c(\mX^{k}_{\ermO}) \right) \right]  \right\} \nonumber \\ 
    & := l_c \left\{ h_c \left( \mX^{j}_{\ermO} \right), f_{\theta} \left[ h_c(\mX^{k}_{\ermO}) \right]  \right\} \quad \forall j,k \in [n],
\end{align}
where $l_c$ is some loss metric between two complete vectors in $\mathbb{R}^{d}$.  

\end{definition}

By Definition \ref{def:1}, we propose to define the cost $c(\cdot, \cdot)$ between two incomplete data points via a distance between their pseudo-completions from $h_c$. As an initializer, $h_c$ can be any off-the-shelf imputer or a measurable function with learnable parameters to be optimized via an objective. The construction of an imputer $h_c$ to create pseudo-complete values is a natural strategy to deal with missing data in this line of research. \citet{muzellec2020missing} and \citet{zhao2023transformed} in particular propose a strategy for learning the imputations by minimizing the OT distance between two pseudo-complete batch of data points. In these works, the imputations for the missing entries are themselves parameters and $h_c$ can be chosen to be such a mechanism. 

Let $\mu_{\theta}(\mX) := n^{-1} \sum^{n}_{j=1} \delta_{f_{\theta}(\mX^{j})}$ denote the model distribution over the true data. 

\begin{lemma}\label{lemma:1} For $h_c, h_m$ defined as above, if $h_c$ is optimal in the sense that $h_c$ recovers the original data i.e., $h_c(\mX^{j}_{\ermO}) = \mX^{j}, \forall j \in [n]$, we have
\begin{align*} 
     W_c \left[\mu_{\gD}(\mX_{\ermO}), \mu_{\theta}(\mX_{\ermO}) \right] = W_{l_c} \left[h_c\#\mu_{\gD}(\mX_{\ermO}), \mu_{\theta}(\mX) \right].
\end{align*}
where $h_c\#\mu_{\gD}(\mX_{\ermO}) = n^{-1} \sum^{n}_{j=1} \delta_{h_c(\mX^{j}_{\ermO})}$, which also represents empirical distribution over the true data. 
\end{lemma}
\textit{Proof.} See Appendix \ref{sup:proof}.

Lemma \ref{lemma:1} follows directly from Definition \ref{def:1}, which translates the problem of minimizing the OT distance between two distributions over the \textit{incomplete} data into minimizing the OT distance between two distributions over the \textit{complete} data produced by $h_c$. Furthermore, if $h_c$ optimally recovers the true data, the minimizer of the OP in (\ref{eq:initial}) coincides with the solution $\theta$ ``best'' fits the true empirical data distribution. 

For brevity, let us denote $\xi$ as the parameters of $h_c$ and alternatively define an empirical distribution $\mu_{\xi}(\widetilde{\mX}) = n^{-1} \sum^{n}_{j=1} \delta_{\widetilde{\mX}^{j}}$ over the set of samples $\left\{\widetilde{\mX}^{j} = h_c(\mX^{j}_{\ermO})\right\}^{n}_{j=1}$. Lemma \ref{lemma:1} motivates an OP for learning $\theta$ by minimizing $W_{l_c} \left[ \mu_{\xi}(\widetilde{\mX}); \mu_{\theta}(\mX)\right]$, where $\mu_{\theta}(\mX)$ is again the model distribution over the true data. On one hand, we cannot compute this OT distance directly because $\mX$ is unknown. On the other hand, Eq. (\ref{eq:dist}) suggests that one can minimize $W_{l_c} \left[ \mu_{\xi}(\widetilde{\mX}); f\#\mu_{\xi}(\widetilde{\mX})\right]$ since fitting $\theta$ to $\widetilde{\mX}$ is the same as fitting $\theta$ to true data $\mX$ if $h_c$ is optimal. However, this is rarely the case in practice because (1) from \S\ref{sect:intro}, we learn that the existing imputation methods are sub-optimal for learning the true causal model; (2) if $h_c$ is learnable, that $h_c$ is arbitrarily initialized provides no useful information about the true data or model distribution for $\theta$ to be updated effectively over training.

We now present our theoretical contributions that open the door to solving the prevailing problem via a more sensible optimization objective. 

\begin{theorem}\label{theorem:1} Given any complete data distribution $\mu_{\xi}(\widetilde{\mX})$, let $\phi: \mathbb{R}^{n \times d} \mapsto \mathbb{R}^{n \times d}$ be a stochastic map such that $\phi\#\mu_{\xi}(\widetilde{\mX}) = \mu_{\theta}(\mX)$ and $h_m \left[ \phi(\widetilde{\mX})\right] = \mX_{\ermO}$. For a fixed value of $\theta$, 
\begin{align}\label{eq:theorem}
&W_{l_c} \left[ \mu_{\xi}(\widetilde{\mX}); \mu_{\theta}(\mX) \right] 
 \nonumber\\ 
& = \min_{\phi} \ \mathbb{E}_{
\widetilde{\mX} \sim \mu_{\xi}(\widetilde{\mX}),
\mY \sim \phi(\widetilde{\mX})
}
\left[ l_c \left( \widetilde{\mX}, f_{\theta}(\mY) \right) \right],
\end{align}
where $l_c$ is a loss metric between two continuous vectors. 
\end{theorem}
\textit{Proof.} See Appendix \ref{sup:proof}.

The push-forward operation $\phi\#\mu_{\xi}(\widetilde{\mX}) = \mu_{\theta}(\mX)$ dictates the existence of a function $\phi$ that transports the data distribution $\mu_{\xi}(\widetilde{\mX})$ to the model distribution such that the OT distance $W_{l_c} \left[ \mu_{\xi}(\widetilde{\mX}); \mu_{\theta}(\mX)\right]$ is tractable for any initializer $h_c$. The second condition $h_m \left[ \phi(\widetilde{\mX})\right] = \mX_{\ermO}$ is to restrict $\phi$ to the class of functions that preserves the observed part of the data. To explain Theorem \ref{theorem:1}, let us recall the data generative mechanism according to an SCM, where a realization from $\mu_{\theta}(\mX)$ is produced by first obtaining values for the root nodes and then generating the data for the remaining nodes via ancestral sampling. Furthermore, if a sample $\mX^{j}$ is indeed generated by the model, we should be able to reconstruct $\mX^{j}$ via the mechanisms $f_{\theta}$. At a certain $\theta$, the optimal $\phi$ ensures the available data is actually generated from the SCM induced by $\theta$. If $\phi$ is at optimality, optimizing (\ref{eq:theorem}) is equivalent to finding $\theta$ that can reconstruct the observed part of the data with minimal cost. 

We first explain how to optimize for $\phi$ and elaborate on the intuition behind its construction in the next section. For fixed $\theta$, the push-forward constraint ensures every sample $\mY^{j} \sim \phi(\widetilde{\mX}^{j})$ to be generated according to the current SCM or more concretely, from the value of the root node in the current causal graph. Note that from one root data value, the model can generate many different samples sharing the same joint distribution. Optimizing mapping $\phi$ forces $\mY$ to have the same distribution with $f_{\theta}(\mY)$ at the current $\theta$. 

Let $\mu_{\phi}(\mY) := \mathbb{E}_{\widetilde{\mX}} \left[ \phi(\mY \vert \widetilde{\mX}) \right]$ be the marginal distribution induced by $\phi$. By the above logic, we relax the constraint into minimizing some divergence between $\mu_{\phi}$ and $f\#\mu_{\phi}$ where $f\#\mu_{\phi}(\mY) :=  n^{-1} \sum^{n}_{j=1} \delta_{f_{\theta}(\mY^{j})}$ for $\mY^{j} \sim \mu_{\theta}(\mY)$. We can then make our OP unconstrained by adding this penalty to (\ref{eq:theorem}). 

Considering $\xi$ to be learnable parameters of $h_c$, our \textbf{final optimization objective} is given by
\begin{align}\label{eq:final}
& \gL(\xi; \theta; \phi) = \min_{\xi,\theta,\phi} \ \mathbb{E}_{
    \widetilde{\mX} \sim \mu_{\xi}(\widetilde{\mX}),
    \mY \sim \phi(\widetilde{\mX})
    } 
    \left[ l_c \left( \widetilde{\mX}, f_{\theta}(\mY) \right) \right] \nonumber\\   
 &  + \lambda \ D \big(\mu_{\phi}; f\#\mu_{\phi}\big) + \gamma_1 \ R(\rmW) + \gamma_2 \ \Vert \rmW \Vert_{1},
\end{align}
where $\lambda, \gamma_1, \gamma_2 > 0$ are trade-off hyper-parameters;  $R(\rmW)$ is any acyclicity regularizer of choice and $\Vert \rmW \Vert_{1}$ is to encourage sparsity; $D$ is an arbitrary divergence measure which can be optimized with maximum mean discrepancy $\textsf{(MMD)}$, the Wasserstein distance or adversarial training. 

\begin{figure}[th!]
    \centering
    \includegraphics[width=\linewidth]{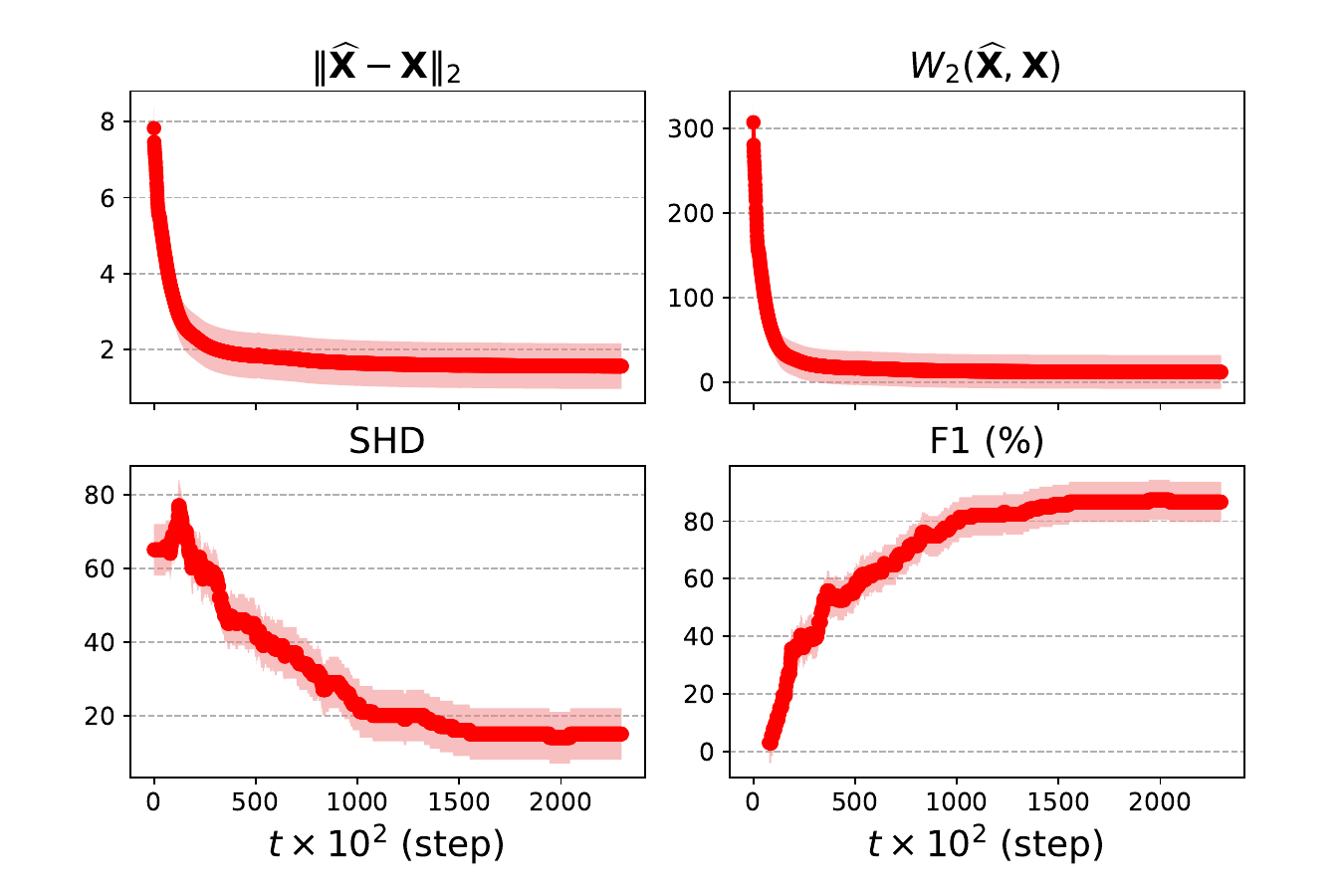}
    \vspace{-3mm}
    \caption{Visualization of the optimization process of OTM. $\rmX$ is the ground-true complete data. $\widehat{\rmX} = f_{\theta}\left[\phi(\rmX_{\ermO}) \right]$ is estimated complete data from the model. 
    \textbf{Top:} As training progresses, the model generates imputations that are closer to the original data both by Euclidean distance (value-wise) and Wasserstein distance (distribution-wise). \textbf{Bottom:} The quality of the estimated graph improves accordingly over training.}
    \label{fig:quali}
\end{figure}

\paragraph{How OTM works.} We now explain the intuition behind the objective in (\ref{eq:final}). Theoretically, the key role of $\phi$ is to render a tractable formulation for our desired OT distance in (\ref{eq:theorem}) when it is difficult to obtain samples from $\mu_{\theta}(\mX)$. To understand how $\phi$ supports optimization, we can view $\phi$ as a ``correction" network that rectifies the random imputations made by $h_c$. Lemma \ref{lemma:1} implies that the closer $h_c$ is to optimality, the better the imputed data approximates the true data, which in turn helps the learning of the graph. Conversely, the fact that $\theta$ optimally yields a distribution, from which the true data is actually generated, would also support the optimization of $h_c$. To facilitate this desired dynamic between $\phi$ and $\theta$,  $\phi$ pushes $h_c$ closer to optimality by concretely forcing the imputed data to obey the dependencies given by the current SCM. This would result from the effect of the second term in (\ref{eq:final}). Without this \textit{distribution matching} term, intuitively, optimizing $\theta$ alone on the first term (for the element-wise reconstruction) would be meaningless since the output imputation from $h_c$ is often non-informative. Our empirical evidence suggests that such a dynamic takes place in our OTM framework where $\phi$ and $\theta$ supports one another to converge to a solution closest to the ground-truth. To further understand how OTM works, we visualize the training behavior in Figure \ref{fig:quali}. By minimizing the OT cost in (\ref{eq:theorem}), OTM seeks to find an SCM that generates data that are ``most similar'' to the original data both in terms of reconstruction (Euclidean distance) and distribution (Wasserstein distance). The figure shows that the causal structure resulting from such an SCM indeed approaches the ground-true one. The objective (\ref{eq:final}) facilitates a joint optimization procedure over both $\phi$ and $\theta$, which can be solved with gradient-based methods. It is worth noting that our formulation so far has not assumed any particular form for the functions $f$.

\paragraph{Implementation details.} The objective (\ref{eq:final}) applies to any DAG characterizations. The integration can simply be done by replacing the regularizer $R(\rmW)$ in Eq. (\ref{eq:final}) with the appropriate formulation of choice.  This means that our algorithm can accommodate any existing score-based causal discovery algorithms with complete data. In the non-linear experiments, we choose DAGMA \citep{bello2022dagma} as our base causal discovery framework, which supports efficient optimization with Adam optimizers \citep{kingma2014adam}. 

Given the optimal $\phi$, our OP reduces to score-based causal discovery on the complete data, where the loss $l_c$ coincides with the score function. We choose $l_c$ to be the MSE-based loss following the default implementation of DAGMA. As for the divergence measure,  one could choose $D$ to be Jensen–Shannon divergence and estimate it with adversarial training \citep{goodfellow2020generative}. However, it is well known that $\textsf{GAN}$ is fairly unstable, which also makes our framework more complex by introducing an additional discriminator. Another consideration is maximum mean discrepancy $(\textsf{MMD})$, which can be estimated empirically with finite samples and whose sample complexity does not depend on the intrinsic dimensionality of the support \citep{sriperumbudur2012empirical}. See Eq. (\ref{eq:mmd}) in the Appendix for the formulation of $\textsf{MMD}$ with RBF kernel.

For the purpose of computational convenience, we approximate $\mX_{\ermO} \approx \left[\mX^{j}_{i} \text{ if } \mM^{j}_{i} = 0 \text{ else } 0, i \in [d], j \in [n]\right] : \mX_{\ermO} \in \mathbb{R}^{n \times d}$ by filling out zeros at the missing entries. Since both functions $h_c$ and $\phi$ play the role of an missing value imputer, using amortized optimization  \citep{amos2023tutorial}, we can combine them into a ``super'' imputer (denoted as $\Phi$) and model it with a sufficiently expressive deep neural network that acts globally on the entire dataset. We therefore construct the function $\Phi: \mathbb{R}^{n \times d} \mapsto \mathbb{R}^{n \times d}$ as $\Phi(\mX_{\ermO}) = \textsf{NN} (\mX_{\ermO}) \circ \mM + \mX_{\ermO}$, where $\textsf{NN}(\cdot)$ is a neural network that returns a distribution over complete data values given $\mX_{\ermO}$. Following the definition of $\phi$, here only the missing entries in the data matrix are filled while the values at the observed entries are retained. We note that OTM can be applied to various data types, thus the design of $\textsf{NN}(\cdot)$ is fairly flexible. As our experiments mostly deal with continuous variables, we amortize $\textsf{NN}(\cdot)$ with a Gaussian distribution with learnable mean and diagonal covariance matrix. The training procedure of OTM is summarized in Algorithm \ref{algo}. 

\paragraph{Identifiability.} One key challenge of learning probabilistic models under missing data is its identifiability. It is worth noting again that our paper considers non-linear ANMs where the causal graph is uniquely identifiable from observed distribution. It is a well-known result that in the M(C)AR cases, the missing mechanism is ignorable and the parameters can be consistently estimated based only on the observed data under large sample assumption \citep{rubin1976inference,mohan2013graphical,bhattacharya2020identification,nabi2020full}. We here provide another perspective on OTM in connection with this result. 

Let $\mX_{\ermM}$ denote that missing part of the missing part of the data. Given the identified parameters $\hat{\theta}$, the complete distribution is recoverable by $\mu_{\hat{\theta}}(\mX) := \mu_{\hat{\theta}}(\mX_{\ermO}, \mX_{\ermM}) = \mu_{\hat{\theta}}(\mX_{\ermM} \vert \mX_{\ermO}, \mM=0)\mu(\mX_{\ermO}, \mM=0)$. The joint recoverability of both the parameters and the complete data distribution is characterized into our optimization routine via the divergence term in Eq. (\ref{eq:final}). The imputation network $\phi$ is essentially optimized such that $\mu_{\phi}(\mX) := \mathbb{E}_{\mX_{\ermO}, \mM=0}\left[\phi(\mX_{\ermO}, \mX_{\ermM} | \mX_{\ermO}, \mM=0)\right] = \mu_{\theta}(\mX)$ at fixed model parameters $\theta$.  
Given sufficient observed data and network capacity, $\mu_{\phi}(\mX)$ can well approximate $\mu_{\theta}(\mX)$ at optimality. Accordingly, as $\theta$ converges to the ground-true solution $\theta^{*}$, $\mu_{\phi}(\mX)$ is expected to fully recover the complete data distribution. 

In the case of MNAR, the missing data mechanism however cannot be ignored during learning. MNAR data remains largely non-recoverable. Existing works  utilize the missingness graph to establish graphical conditions for recovering the joint distribution \citep{bhattacharya2020identification, nabi2020full,ma2021identifiable}. 
The knowledge of the missingness structure is thus necessary, which is however practically hard to obtain, especially in our setting where the causal structure is also unknown. We thus find it crucial to develop a principled approach that operates effectively in the absence of knowledge about the underlying missingness mechanism. We will later provide extensive empirical evidence to  justify the consistent effectiveness of OTM in MNAR cases.

\begin{algorithm}[hbt!]
    \caption{OTM Algorithm}
    \label{algo}
\begin{algorithmic}
\STATE \textbf{Input:} Incomplete data matrix $\mX = \big[\mX^{1}, \cdots, \mX^{n}\big]^T \in \mathbb{R}^{n\times d}$; missing mask $\mM$, regularization coefficients $\lambda, \gamma_1, \gamma_2 > 0$; loss function $l_c$; characteristic positive-definite kernel $\kappa$ and $M-$matrix domain $s > 0$.

\STATE \textbf{Output:} Weighted adjacency matrix $\rmW \in \theta$.

\STATE Initialize the parameters $\theta, \Phi$.

\STATE \textbf{while} $(\Phi, \theta)$ \textit{not converged} \textbf{do}
    
    \STATE \ \ \ \ Set $\mX_{\ermO} = \left[\mX^{j}_{i} \text{ if } \mM^{j}_{i} = 0 \text{ else } 0, i \in [d], j \in [n]\right]$;
    
    \STATE \ \ \ \  Sample $\widetilde{\mX}$ from $\Phi \left( \mX_{\ermO} \right)$;

    \STATE \ \ \ \ Evaluate $\mY = f_{\theta}(\widetilde{\mX})$;

    \STATE \ \ \ \ Update $\Phi, \theta$ by descending 
    \begin{align*}
    & \gL(\Phi, \theta) = \frac{1}{n} \sum^{n}_{j=1} l_c \big(
    \widetilde{\mX}^{j}, \mY^{j} \big) + \lambda \ \textsf{MMD}(\widetilde{\mX}, \mY, \kappa) \\
    & + \gamma_1 \left[ - \log \det \big(s\rmI - \rmW \circ \rmW \big) + d \log s \right] + \gamma_2 \ \Vert \rmW \Vert_{1}.
    \end{align*}
\STATE \textbf{end while}
\end{algorithmic}
\end{algorithm}

\section{Experiments}
We evaluate OTM\footnote{The code is released at \url{https://github.com/isVy08/OTM}.} on both synthetic and real-world datasets. The detailed implementations of all methods are presented in Appendix \ref{sup:config}. We focus on the non-linear setting in the main text since it is more challenging. Additional results on the linear models are reported in Appendix \ref{sup:linear}.

\begin{figure*}[h!]
    \centering
    \includegraphics[width=\linewidth]{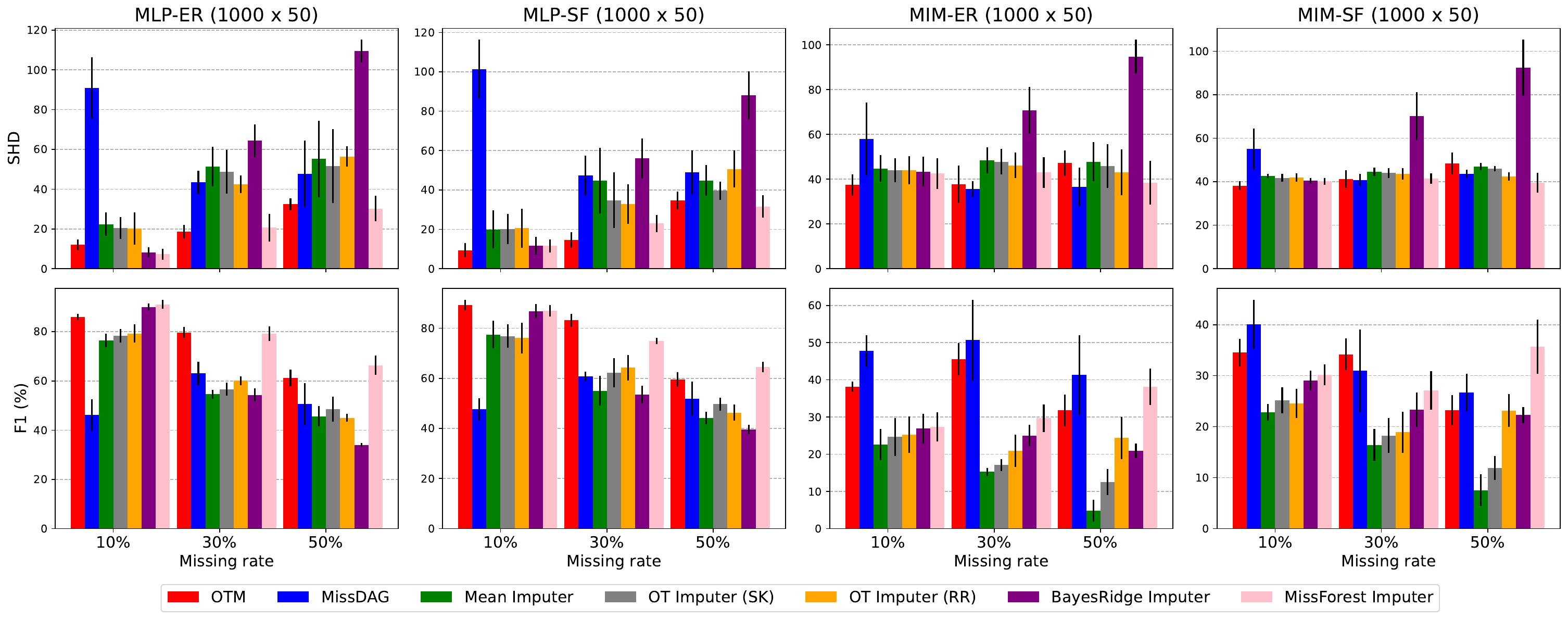}
    \vspace{-3mm}
    \caption{Nonlinear ANMs (MCAR).  \textsf{SK} refers to batch Sinkhorn imputation and \textsf{RR} refers to round-robin Sinkhorn imputation. \\ SHD $\downarrow$ and F1 $\uparrow$.}
    \label{fig:sim-mcar}
\end{figure*}

\begin{figure*}[htb!]
    \centering
    \includegraphics[width=\linewidth]{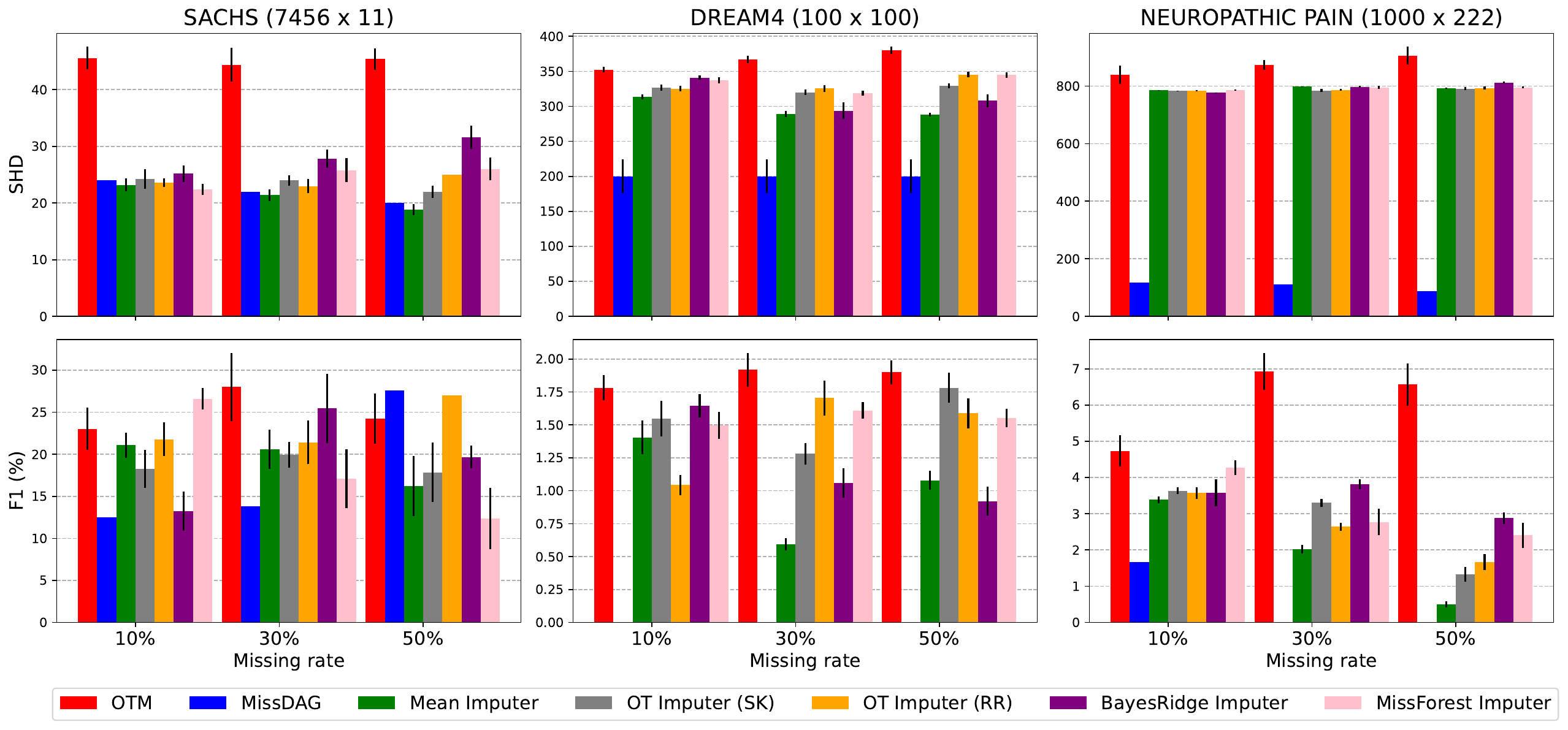}
    \vspace{-3mm}
    \caption{Real-world datasets (MCAR). \textsf{SK} refers to batch Sinkhorn imputation and \textsf{RR} refers to round-robin Sinkhorn imputation. \\ SHD $\downarrow$ and F1 $\uparrow$.}
    \label{fig:real-mcar}
\end{figure*}

\paragraph{Baselines.}
We consider imputation methods as baselines, including Mean imputation, Multivariate imputation: Bayes Ridge regression and Random Forest regression (MissForest) \citep{van2000multivariate,pedregosa2011scikit}, and OT imputation (Batch Sinkhorn and Round-Robin Sinkhorn) \citep{muzellec2020missing} to fill in the missing data and then apply DAGMA \citep{bello2022dagma} for structure learning. We further compare OTM with MissDAG \citep{gao2022missdag} as the key competing method, which has already been shown to outperform many constraint-based baselines. We do not report results for VISL \citep{morales2022simultaneous} as their code is not available to our knowledge.

\paragraph{Thresholding \& Metrics.}
Following the standard causal discovery practice \citep{zheng2018dags,ng2020role}, a final threshold of $0.3$ is applied post-training to ensure the DAG output: it iteratively  removes the edge with the minimum magnitude until the final graph is a DAG. All quantitative results are averaged over $5$ random initializations. For comparing the estimated DAG with the ground-truth one, we report the commonly used metrics: F$1$-score and Structural Hamming Distance (SHD) with SHD referring to the smallest number of edge additions, deletions, and reversals required to transform the recovered DAG into the true one. Lower SHD is preferred ($\downarrow$) while higher F$1$ is preferred ($\uparrow$).  Note that SHD is a standard metric used across causal discovery literature, but it can be biased. Because SHD quantifies the number of errors in absolute value, given a sparse graph, a method could achieve low SHD by predicting few edges, which obviously would compromise the accuracy score. Therefore, one need to examine both metrics to assess the causal discovery performance thoroughly.
  
\subsection{Simulations}
We simulate synthetic datasets generating a ground-true DAG from one of the two graph models, Erdos-Rényi (ER) or Scale-Free (SF). Each function $f_i$ is constructed from a multi-layer perceptron (MLP) and a multiple index model (MIM) with random coefficients. We consider a general scenario of non-equal variances, sampling $1000$ observations according to all missing mechanisms: MCAR, MAR and MNAR at $10\%, 30\%, 50\%$ missing rates. In the main text, we report the results in MCAR cases with the standard setting of $50$ nodes and $100$ directed edges, using Gaussian noise. More results on MAR, MNAR cases as well as experiments with different graph degrees, noise distributions and acyclicity constraints are reported in Appendix \ref{sup:ablation}.

\paragraph{Results.}\label{sect:sim}
The effectiveness of OTM across various settings is demonstrated in Figures \ref{fig:sim-mcar}, \ref{fig:sim-mar} and \ref{fig:sim-mnar}. OTM achieves consistently low SHD scores with the highest/second-highest F$1$-scores. Sub-optimality is again observed in the simulations when a causal discovery method is applied on top of existing imputation baselines. Throughout our simulations, MissForest imputer produces the best imputation quality and works best among the imputers. However, the method later exhibits inconsistencies in its efficacy over the other imputation baselines in real-world settings. This again demonstrates that naively applying causal discovery on the post-imputed data tends to yield unreliable results. 

On the other hand, MissDAG does not perform well in our non-linear settings although it has good performance on linear settings (See Figures \ref{fig:linear}). The key challenge arises from the intractability of the posterior distribution and non-trivial computation of likelihood expectation. MissDAG resorts to an approximate posterior, which requires rejection sampling to fill in the missing values in the E steps. Our empirical evidence reveals that MissDAG still performs adequately in the simulated settings with possibly minor misspecifications. However, as we transition to real-world cases, the challenges become more pronounced, highlighting the limitations of the sampling process in practical scenarios. 

\subsection{Real-world datasets}\label{sect:real}

We evaluate OTM on $3$ well-known biological datasets with ground-true causal relations. The first one is the Neuropathic Pain dataset \citep{tu2019causal}, containing
diagnosis records of neuropathic pain patients. There are $1000$ samples and $222$ binary variables, indicating existence of the symptoms. The second causal graph named Sachs \citep{sachs2005causal} models a network of cellular signals, consisting of $11$ continuous variables and $7466$ samples. Dream4 \citep{greenfield2010dream4,marbach2010revealing} is the last dataset which simulates gene expression measurements from five transcriptional regulator sub-networks of E. coli and S. cerevisiae. The provided data contains $5$ subsets, each has $100$ continuous variables and $100$ samples.  

\begin{figure}[thb!]
    \centering
    \includegraphics[width=\linewidth]{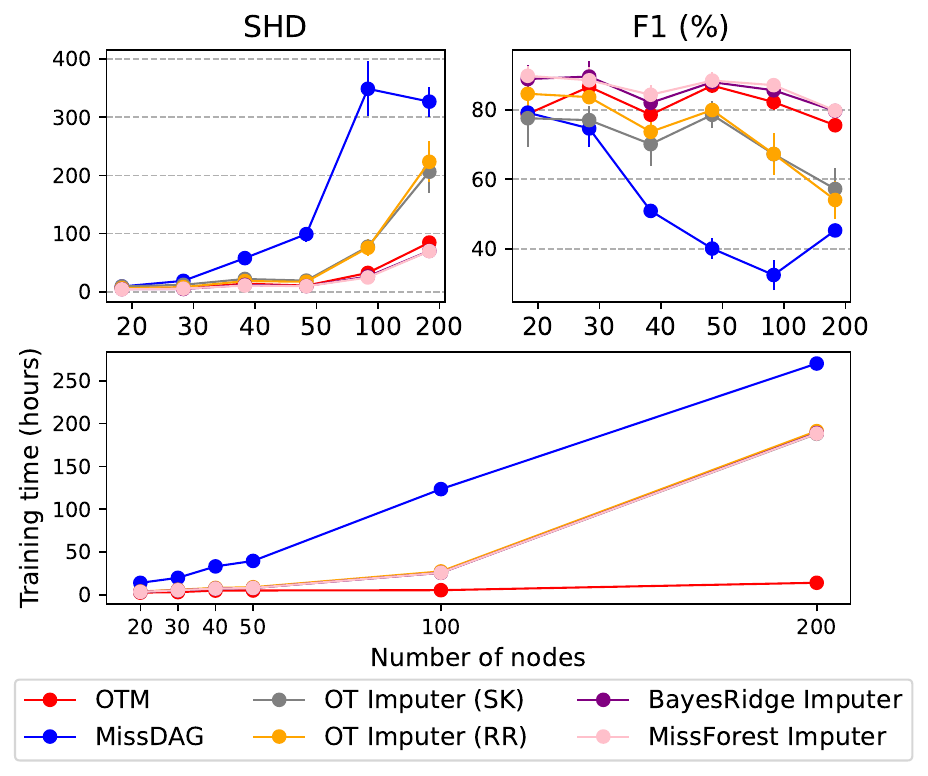}
    \vspace{-3mm}
    \caption{Scalability of methods in nonlinear ANMs (MCAR) at 10\% missing rate. \textsf{SK} refers to batch Sinkhorn imputation and \textsf{RR} refers to round-robin Sinkhorn imputation. SHD $\downarrow$ and F1 $\uparrow$. The training time of the imputation baselines includes the time for learning imputations.}
    \label{fig:scalability}
\end{figure}

\paragraph{Results.}

It is worth noting that the performance of all methods is comparatively lower compared with that in the simulated settings. This is because in real-world scenarios, the model is prone to misspecification and the possible existence of latent confounders. However, it is observed from Figures \ref{fig:real-mcar}, \ref{fig:real-mar} and \ref{fig:real-mnar} that OTM stands out from the baselines with superior accuracy. Meanwhile, MissDAG fails to detect the true edge directions in several settings on Neuropathic pain and Dream4 datasets. In these experiments, we find that its underlying causal discovery method, NOTEARS, does not converge in every EM iteration on the samples obtained from approximate posterior. 

An additional limitation of MissDAG pertains to the scalability of the method. As illustrated in Figure \ref{fig:scalability}, our approach exhibits faster run-time compared to MissDAG. This enhanced efficiency can be attributed to two key differences between OTM and MissDAG. First, OTM runs on DAGMA, the superior scalability of which over NOTEARS has been established in \citet{bello2022dagma}. 
Second, the iterative nature of EM and the incorporation of rejection sampling in the E step contribute to the overall computational burden of MissDAG. Figure \ref{fig:scalability} demonstrates when the number of nodes increases, the performance of MissDAG degrades significantly and the training of MissDAG becomes more computational expensive. Such a sub-optimal behavior also translates to the real-world settings, where MissDAG specifically fails on Neuropathic Pain and Dream4 datasets of high dimensionality.

\section{Limitations and Future Works} 
We have proposed OTM, a framework based on optimal transport for DAG learning under missing data. OTM is shown to be more robust and scalable approach while being flexible  to accommodate many existing score-based causal discovery algorithms. Our training routine is strongly motivated by the principles for identifying the model parameters and complete data distribution under M(C)AR data in previous literature. However, since our optimization objective with the DAG constraint is highly non-convex, it is thus difficult to provide theoretical convergence results, which  remains a challenge for optimization-based causal discovery algorithms in non-linear ANMs \citep{wei2020dags,ng2022convergence,gao2022missdag}. Furthermore, due to its prevalence in real-world settings, identifiability of the causal graph under MNAR data is a significant research problem that deserves further efforts. Given the effectiveness of our current approach, we believe leveraging extra graphical conditions about MNAR data would help improve the performance. Other future directions include more challenging settings wherein OTM is currently inapplicable, such as cyclic causal graphs or graphs with latent confounders.  

\section*{Acknowledgments}
Trung Le and Dinh Phung were supported by ARC DP23 grant DP230101176 and by the Air Force Office of Scientific Research under award number FA2386-23-1-4044. This does not imply endorsement by the funding agency of the research findings or conclusions. Any errors or misinterpretations in this paper are the sole responsibility of the authors.

\section*{Impact Statement}
This work introduces an application of machine learning to effectively address a class of statistical estimation problems in a scalable manner. While we are currently unaware of any potential negative societal impacts of our work, machine learning frequently yields unintended consequences in various domains, necessitating thorough consideration of societal advantages and drawbacks when implementing the proposed method in real-world scenarios.

\bibliographystyle{icml2024}
\bibliography{ref}
\newpage
\appendix
\onecolumn
\section{Proofs} \label{sup:proof}
To make the proof self-contained, we here recap the problem setup in the main text. 

Let $\mX$ be the true (unknown) data matrix and $\mX_{\ermO}$ denote the observed part of the data that is available. Given a complete matrix $\mX$, we define $h_m$ as a mechanism for producing the missing data, or equivalently extracting the observed part i.e., $h_m(\mX) = \mX_{\ermO}$ and $h_m$ operates row-wise according to a given missing mask $\mM$. Given an observational instance $\mX^{j}_{\ermO}$, we construct a mechanism $h_c$ to complete the data such that $h_c(\mX^{j}_{\ermO}) = \widetilde{\mX}^{j} \in \mathbb{R}^{d}$.  

For $h_m$ and $h_c$ defined above, we can define the model distribution over $\mX_{\ermO}$ via its reconstruction from the model, that is $\mu_{\theta}(\mX_{\ermO}) := n^{-1} \sum^{n}_{j=1} \delta_{h_m \left\{ f_{\theta} \left[ h_c(\mX^{j}_{\ermO}) \right] \right\} }$. Furthermore, let $\mu_{\theta}(\mX) := n^{-1} \sum^{n}_{j=1} \delta_{f_{\theta}(\mX^{j})}$ denote the model distribution over the true data.

\textbf{Definition \ref{def:1}.} (Cost function for incomplete samples) The transport cost between a particle $j$ from $\mu_{\gD}(\mX_{\ermO})$ and a particle $k$ from $\mu_{\theta}(\mX_{\ermO})$ is given by
    \begin{align}\label{eq:distance}
    & c \left\{ \mX^{j}_{\ermO}, h_m \left[ f_{\theta} \left( h_c(\mX^{k}_{\ermO}) \right) \right]  \right\} \nonumber 
    := l_c \left\{ h_c \left( \mX^{j}_{\ermO} \right), f_{\theta} \left[ h_c(\mX^{k}_{\ermO}) \right]  \right\} \quad \forall j,k \in [n],
\end{align}
where $l_c$ is a metric between two complete vectors in $\mathbb{R}^{d}$.

\subsection{Proof for Lemma \ref{lemma:1}.}
\textbf{Lemma \ref{lemma:1}.} For $h_c, h_m$ defined as above, if $h_c$ is optimal in the sense that $h_c$ recovers the original data i.e., $h_c(\mX^{j}_{\ermO}) = \mX^{j}, \forall j \in [n]$, we have
\begin{align*} 
     W_c \left[\mu_{\gD}(\mX_{\ermO}), \mu_{\theta}(\mX_{\ermO}) \right] = W_{l_c} \left[h_c\#\mu_{\gD}(\mX_{\ermO}), \mu_{\theta}(\mX) \right].
\end{align*}
where $h_c\#\mu_{\gD}(\mX_{\ermO}) = n^{-1} \sum^{n}_{j=1} \delta_{h_c(\mX^{j}_{\ermO})}$, which also represents empirical distribution over the true data.

\begin{proof}
   \begin{align*}
    & W_c \left[\mu_{\gD}(\mX_{\ermO}), \mu_{\theta}(\mX_\ermO) \right] \nonumber \\ 
    & = \min_{\rmP \in \sU(\mathbf{1}_{n},\mathbf{1}_{n}) } \sum_{j,k}  c \left\{ \mX^{j}_{\ermO}, h_m \left[ f_{\theta} \left( h_c(\mX^{k}_{\ermO}) \right) \right]  \right\} \rmP^{j,k} \\
    & \overset{(1)}{=} \min_{\rmP \in \sU(\mathbf{1}_{n},\mathbf{1}_{n}) } \sum_{j,k} l_c \left\{ h_c \left( \mX^{j}_{\ermO} \right), f_{\theta} \left[ h_c(\mX^{k}_{\ermO}) \right]  \right\} \rmP^{j,k} \\
    & \overset{(2)}{=} \min_{\rmP \in \sU(\mathbf{1}_{n},\mathbf{1}_{n}) } \sum_{j,k} l_c \left\{ h_c(\mX^{j}_{\ermO}) ,  f_{\theta}\left( \mX^{k} \right)  \right\}\rmP^{j,k} \\ 
    & = W_{l_c} \left[h_c\#\mu_{\gD}(\mX_{\ermO}), \mu_{\theta}(\mX) \right],
\end{align*}
where the minimum is taken over all possible couplings in the Birkhoff polytope $\sU(\mathbf{1}_{n}/n,\mathbf{1}_{n}/n)$. Note that the equality $\overset{(1)}{=}$ follows from Definition \ref{def:1}. We further have $\overset{(2)}{=}$ since $h_c$ is optimal. 
\end{proof}

\subsection{Proof for Theorem \ref{theorem:1}.}
Given a set of observed samples $\mX_{\ermO} = \{\mX^{j}_{\ermO}\}_{j=1}^{n}$ and an imputer $h_c$ with parameters $\xi$, let $\mu_{\xi}(\widetilde{\mX}) = n^{-1} \sum^{n}_{j=1} \delta_{\widetilde{\mX}^{j}}$ define an empirical distribution  over the set of samples $\left\{\widetilde{\mX}^{j} = h_c(\mX^{j}_{\ermO})\right\}^{n}_{j=1}$. 

\textbf{Theorem \ref{theorem:1}.} Given any complete data distribution $\mu_{\xi}(\widetilde{\mX})$, let $\phi: \mathbb{R}^{n \times d} \mapsto \mathbb{R}^{n \times d}$ be a stochastic map such that $\phi\#\mu_{\xi}(\widetilde{\mX}) = \mu_{\theta}(\mX)$ and $h_m \left[ \phi(\widetilde{\mX})\right] = \mX_{\ermO}$. For a fixed value of $\theta$, 
\begin{align}
&W_{l_c} \left[ \mu_{\xi}(\widetilde{\mX}); \mu_{\theta}(\mX) \right] = \min_{\phi} \ \mathbb{E}_{
\widetilde{\mX} \sim \mu_{\xi}(\widetilde{\mX}),
\mY \sim \phi(\widetilde{\mX})
}
\left[ l_c \left( \widetilde{\mX}, f_{\theta}(\mY) \right) \right],
\end{align}
where $l_c$ is a metric between two complete vectors in $\mathbb{R}^{d}$.

\begin{proof} 
A sample $\mX^{j} \sim \mu_{\theta}(\mX)$ is realized from an SCM by first sampling from the model the values for the root nodes and then generating the data for the remaining nodes via ancestral sampling. If a sample $\mX^{j}$ is indeed generated by the model, we should be able to reconstruct $\mX^{j}$ via the mechanisms $f_{\theta}$, where every feature $\mX^{j}_{i}$ is only determined by the features corresponding to the parent nodes of node $i$, while the effect of the non-parental features is zero-ed out by $\rmW(f)$. 

In practical implementation, amortized optimization is often employed where $f_{\theta}$ is parameterized with a single, yet deep neural network. $f_{\theta}$ is a sufficiently expressive function that can factor in the noise effect. In light of the insight, one thus can relax the setting by viewing $\mX$ as its own parents and $\mX$ reconstructs itself deterministically i.e., $\mX = f_{\theta}(\mX)$.   

We consider three distributions: $\mu_{\xi}(\widetilde{\mX})$ over $A = \mathbb{R}^{n \times d}$, $\mu_{\theta}(\mY)$ over $B = \mathbb{R}^{n \times d}$ and $\mu_{\theta}(\mX)$ over $C = \mathbb{R}^{n \times d}$. Note that the establishment of $B$ and $C$ over the same distribution follows from the above mechanism where $\mX$ reconstructs itself and we denote the data matrix as $\mY$ to distinguish the roles of the two distributions.   

Let $\Gamma^{*} \in \gP\left(\mu_{\xi}(\widetilde{\mX}), \mu_{\theta}(\mX)\right)$ be the \textit{optimal} joint distribution over the $\mu_{\xi}(\widetilde{\mX})$ and $\mu_{\theta}(\mX)$ of the corresponding Wasserstein distance.  Let $\alpha = (id, f)\#\mu_{\theta}(\mY)$ be a deterministic coupling or joint distribution over $\mu_{\theta}(\mY)$ and $\mu_{\theta}(\mX)$. 

The Gluing lemma  \citep[see Lemma 5.5 in][]{santambrogio2015optimal} indicates the existence of a tensor coupling measure $\sigma$ over $A \times B \times C$ such that $(\pi_{A,C})\#\sigma = \Gamma$ and  $(\pi_{B, C})\#\sigma = \alpha$, where $\pi$ are the projectors. Let $\beta = (\pi_{A,B})\#\sigma$ be a joint distribution over $\mu_{\xi}(\widetilde{\mX})$ and $\mu_{\theta}(\mY)$. 

Let $\phi(\widetilde{\mX}) = \beta(\cdot \vert \widetilde{\mX})$ further denote a stochastic map from $A$ to $B$. Let $\sigma_{BC} = (\pi_{B,C})\#\sigma$.  It follows that 
\begin{align}\label{eq:proof1}
   W_{l_c} \left[ \mu_{\xi}(\widetilde{\mX}); \mu_{\theta}(\mX) \right]  & = \mathbb{E}_{(\widetilde{\mX}, \mX) \sim \Gamma^{*}} \left[l_c (\widetilde{\mX}, \mX) \right] = \mathbb{E}_{(\widetilde{\mX}, \mY, \mX )\sim \sigma} \left[l_c (\widetilde{\mX}, \mX) \right] \nonumber \\
   & = \mathbb{E}_{\widetilde{\mX} \sim \mu_{\xi}({\widetilde{\mX}}), \mY \sim \beta(\cdot \vert \widetilde{\mX}), \mX \sim \sigma_{BC}(\cdot \vert \mY)} \left[l_c (\widetilde{\mX}, \mX) \right] \nonumber \\
   & \overset{(3)}{=} \mathbb{E}_{\widetilde{\mX} \sim \mu_{\xi}({\widetilde{\mX}}), \mY \sim \phi(\widetilde{\mX}), \mX = f_{\theta}(\mY)} \left[l_c (\widetilde{\mX}, \mX) \right] \nonumber \\ 
   & \overset{(4)}{=} \mathbb{E}_{\widetilde{\mX} \sim \mu_{\xi}({\widetilde{\mX}}), \mY \sim \phi(\widetilde{\mX})} \left[l_c \left( \widetilde{\mX}, f_{\theta}(\mY) \right) \right] \nonumber \\
   & \ge \min_{\phi} \quad \mathbb{E}_{\widetilde{\mX} \sim \mu_{\xi}({\widetilde{\mX}}), \mY \sim \phi(\widetilde{\mX})} \left[l_c \left(\widetilde{\mX}, f_{\theta}(\mY) \right) \right]. 
\end{align}

Let $\phi^{*}$ be the \textit{optimal} backward map satisfying $\phi\#\mu_{\xi}(\widetilde{\mX}) = \mu_{\theta}(\mY)$. The joint distribution $\beta$ over $\mu_{\xi}(\widetilde{\mX})$ and $\mu_{\theta}(\mY)$ is now constructed by first sampling $\widetilde{\mX}$ from $\mu_{\xi}(\widetilde{\mX})$, then sampling $\mY$ from $\phi^{*}(\widetilde{\mX})$ and finally collecting $(\widetilde{\mX}, \mY) \sim \beta$. 

Let us consider the joint distribution $\alpha$ over $\mu_{\theta}(\mY)$ and $\mu_{\theta}(\mX)$ as defined above. By the Gluing lemma, there exists a joint distribution $\sigma$ over $A \times B \times C$ such that $(\pi_{A,B})\# \sigma = \beta$ and  $(\pi_{B,C})\# \sigma = \alpha$. We denote $\Gamma = (\pi_{A,C})\#\sigma$ the induced joint distribution over $\mu_{\xi}(\widetilde{\mX})$ and $\mu_{\theta}(\mX)$. Let $\sigma_{BC} = (\pi_{B,C})\#\sigma$.
\begin{align} \label{eq:proof2}
    & \min_{\phi} \quad \mathbb{E}_{\widetilde{\mX} \sim \mu_{\xi}(\widetilde{\mX}), \mY \sim \phi(\widetilde{\mX})} \left[l_c \left( \widetilde{\mX}, f_{\theta}(\mY) \right) \right] \nonumber \\ 
    & = \mathbb{E}_{\widetilde{\mX} \sim \mu_{\xi}(\widetilde{\mX}), \mY \sim \phi^{*}(\widetilde{\mX}) } \left[l_c (\widetilde{\mX}, f_{\theta}(\mY)) \right]\nonumber \\ 
    & \overset{(5)}{=} \mathbb{E}_{\widetilde{\mX} \sim \mu_{\xi}(\widetilde{\mX}), \mY \sim \phi^{*}(\widetilde{\mX}), \mX = f_{\theta}(\mY)} \left[l_c (\widetilde{\mX}, \mX ) \right] \nonumber \\ 
    & = \mathbb{E}_{\widetilde{\mX} \sim \mu_{\xi}(\widetilde{\mX}), \mY \sim \beta(.\vert \widetilde{\mX}), \mX \sim \sigma_{BC}(. \vert \mY)} \left[l_c (\widetilde{\mX}, \mX) \right] \nonumber \\
    & = \mathbb{E}_{(\widetilde{\mX}, \mY, \mX)\sim \sigma} \left[l_c (\widetilde{\mX}, \mX) \right] \nonumber \\
    & = \mathbb{E}_{(\widetilde{\mX}, \mX) \sim \Gamma} \left[l_c (\widetilde{\mX}, \mX) \right] \nonumber \\
    & \ge \min_{\Gamma \sim \gP\left(\mu_{\xi}(\widetilde{\mX}), \mu_{\theta}(\mX)\right)} \mathbb{E}_{(\widetilde{\mX}, \mX) \sim \Gamma} \left[l_c (\widetilde{\mX}, \mX) \right] = W_{l_c} \left[ \mu_{\xi}(\widetilde{\mX}); \mu_{\theta}(\mX)\right].
\end{align}

\end{proof}

Note that the equality's in $(3)-(5)$ are due to the fact that $\alpha$ is a deterministic coupling and the expectation is reserved through a deterministic push-forward map. From (\ref{eq:proof1}) and (\ref{eq:proof2}), we reach the desired equality.

\section{Related Work}\label{sect:rwork}
\paragraph{Missing data imputation.} Besides basic imputation with mean/median/mode values, there are a plethora of advanced methods for filling in the missing values. In one class of methods, data features are handled one at a time. The key technique is to iteratively estimate the conditional distribution of the feature given the other features \cite{van2006fully,van2011mice,liu2014stationary,zhu2015convergence}. Methods in the other class consider features altogether by learning their joint distribution explicitly or implicitly \cite{mattei2019miwae,yoon2020gamin,nazabal2020handling,richardson2020mcflow,you2020handling,dai2021multiple,peis2022missing,fang2023fragmgan,gao2023handling}. A separate line of works focuses on refining existing imputation methods, which show improvements over stand-alone methods \cite{yoon2018gain,mohan2021graphical,jarrett2022hyperimpute}. 

\paragraph{Causal discovery with complete data.} Causal discovery algorithms primarily fall into two categories: constraint-based and score-based approaches.  Constraint-based methods such as PC \cite{spirtes1991algorithm} and FCI \cite{spirtes2000causation} extract conditional independencies from the data distribution to detect edge existence and direction. These approaches have been adapted to address the issue of missing data through test-wise deletion and adjustments \cite{strobl2018fast,gain2018structure,tu2019causal}. On the other hand, score-based methods search for model parameters in the DAG space by optimizing a scoring function \cite{ott2003finding,chickering2002optimal,teyssier2012ordering,cussens2017polyhedral}. Historically, such methods come with significant computational burden due to combinatorial optimization complexities. Continuous optimization of structures, pioneered by NOTEARS \cite{zheng2018dags}, later lays the foundation for the development of scalable causal discovery methods. NOTEARS introduces an algebraic characterization of DAGs via trace exponential, extending its applicability to nonlinear scenarios \cite{lachapelle2019gradient,zhu2019causal,wang2020causal,wang2021ordering,ng2022masked}, time-series data \cite{pamfil2020dynotears}, unmeasured confounders \cite{yuan2011learning}, and topological ordering \cite{deng2023optimizing}. Alternative DAG characterizations also exist such as based on the polynomial formulation \cite{yu2019dag} and log determinant \cite{bello2022dagma}. While early methods focus on the point estimation of graphs, modern works adopt Bayesian approach learning distributions over graphs. These techniques can incorporate DAG formulations seamlessly within their frameworks, capitalizing on the differentiability of DAG sampling and leveraging an amortized inference engine for enhanced efficiency \cite{lorch2021dibs,ashman2022causal,geffner2022deep,lorch2022amortized,charpentier2022differentiable}.

\paragraph{Causal discovery with missing data.} Extensions of the PC algorithm exist for learning causal graphs under missing data \citep{tu2019causal,gain2018structure}, which utilizes all the samples without missingness while   eliminating the biases involved in the conditional independence test. A dominant family of methods is based on Expectation-Maximization \citep[EM,][]{Dempster1977} that iteratively infers missing values and performs structural learning. \citet{adel2017learning} introduce a hill-climbing approximate algorithm for the completions
of the missing values, which is followed by structure optimization step by any off-the-shelf algorithm for structure learning. \citet{friedman1997learning} and \citet{singh1997learning} iteratively refine conditional distributions from which they sample missing values. These methods require the posterior exists in closed form and can only discover the graph up to the Markov equivalence class. Leveraging continuous optimization from NOTEARS \cite{zheng2018dags}, MissDAG \cite{gao2022missdag} focuses on continuous identifiable ANMs and develop a method based on approximate posterior using Monte Carlo and rejection sampling where exact posterior is not available. VISL \cite{morales2022simultaneous} is a divergent line of approach based on amortized variational inference. Different from OTM and MissDAG, VISL adopts Bayesian learning and assumes the existence of a latent, low-dimensional factor that effectively summarizes the data based solely on the observed part. The model learns the latent factors of the data, using which to discover the graph and reconstruct the full data. 

\paragraph{Optimal transport.}
Optimal transport (OT) studies the optimal transportation of mass from one distribution to another \cite{villani2009optimal}. Through the notion of Wasserstein distance, OT offers a geometrically meaningful distance between probability distributions, proving effectiveness in various machine learning domains ~\cite{huynh2020otlda,zhao2020neural,nguyen2021most,wanrepresenting2022,zhang2022deepemd,bui2021unified,nguyen2022cycle,vuong2023vector,ye2024ptarl,gao2024distribution,luong2024revisiting,vo2024parameter}. From the view of distribution matching, OT-based missing data imputation has been proposed \cite{muzellec2020missing,zhao2023transformed}. The key idea is to learn the missing values by minimizing the distribution between two randomly sampled batches in the data. An emerging line of research is OT for causal discovery, initiated by \citet{tu2021optimal}, who provide new identifiability results based on a dynamic formulation of OT. This work considers a simple two-node setting solely for additive noise models. A modern extension is proposed in \citet{akbari2023learning}, yet only applicable to complete data. 


\section{Experiment Details}\label{sup:config}
This section provides the implementation details for OTM and the baseline methods, as well as the mechanism used to produce the missing data. 

\subsection{Training configuration}
We apply the default structure learning setting of DAGMA to OTM and the imputation baselines: log \textsf{MSE} for the non-linear score function, number of iterations $T = 4$, initial central path coefficient $\mu = 1$, decay factor $\alpha = 0.1$, log-det parameter $s = \{1, 0.9, 0.8, 0.7\}$. DAGMA implements an adaptive gradient method using the ADAM optimizer with learning rate of $2 \times 10^{-4}$ and $(\gamma_1, \gamma_2) = (0.99, 0.999)$, where $\gamma_2$ is the coefficient for $l_1$ regularizer included to promote sparsity.

\paragraph{OTM.} As for OTM implementation, we note that OTM can be applied to various data types. As our experiments mostly deal with continuous variables, we model the imputation network $\textsf{NN}(\cdot)$ with a local Gaussian distribution and design two multi-layer perceptrons (with ReLU activation) that output the means and diagonal covariance matrices given the missing data, where the missing entries are pre-imputed with zeros. The hidden units are equal to the number of nodes in the graph. For real-world datasets only, we add an extra Tanh+ReLU activation as the final layer since the input values are between $0$ and $1$. The imputation network is integrated into DAGMA, and all parameters are optimized for $8 \times 10^{3}$ iterations. The optimal values of $\lambda$ are found to be $0.01$ for MLP models (the correctly specified setting) and $1.5$ for the other models (mis-specified settings). For the divergence measure $D$, we use \textsf{MMD} with RBF kernel in our experiments. 

The squared MMD between two set of samples $\mU = \{\mU^{j}\}^{n}_{j=1}$ and $\mV = \{\mV^{j}\}^{n}_{j=1}$, where $\mU^{j}, \mV^{j} \in \mathbb{R}^d$ is computed as  
\begin{align}\label{eq:mmd}
\textsf{MMD}^{2}(\mU, \mV, \kappa) =  \frac{1}{n(n-1)} \sum_{j \ne k} \kappa \big( \mU^{j}, \mU^{k} \big) +
\frac{1}{n(n-1)} \sum_{j \ne k} \kappa \big( \mV^{j}, \mV^{k} \big) - 
\frac{2}{n^2} \sum_{j,k} \kappa \big(\mU^{j}, \mV^{k} \big),
\end{align}
where $\kappa$ is the characteristic positive-definite kernel. 

\paragraph{Baselines.}
For Mean imputation and Iterative imputation (the number of iterations is $50$), we use \texttt{Scikit-learn} implementation. For OT imputer \citep{muzellec2020missing}, we set the learning rate to $0.01$ and the number of iterations to $10,000$. Following their code, we run DAGMA on the imputed data for $8 \times 10^{4}$ iterations.  MissDAG is currently built on NOTEARS \citep{zheng2018dags}. For MissDAG, we set the number of iterations for the EM procedure is $10$ on synthetic datasets and $100$ for real datasets, while leaving the other hyper-parameters of MissDAG same as reported in the paper. 

\subsection{Missingness}
We follow the setup of MissDAG \citep{gao2022missdag} to generate the missing indicator matrix $\mM$ that is used to mask the original data. For completeness, we repeat the procedure here. Across the experiments, we set $30\%$ of the variables to be fully-observed. Let $r_m$ denote the missing rate. 

\begin{itemize}
    \item \textbf{MCAR:} This missing mechanism is independent from the variables. We sample a matrix $\mM'$ from a $\textrm{Uniform}([0,1])$ and set $\mM[t,i]=0$ if $\mM'[t,i] \le r_m$.
    
    \item \textbf{MAR:} The missing mechanism is systematically related to the observational variables but independent from the missing variables. The missingness of the remaining $70\%$ variables are generated according to a logistic model with random weights that are related to the fully observed variables.

    \item \textbf{Logistic MNAR:} The missing mechanism is related to the missing variables. The remaining $70\%$ variables are split into a set of inputs for a logistic model and a set whose missing probabilities are
    determined by the logistic model. Then inputs are then masked MCAR, so the missing values from the second set will
    depend on masked values.
    Weights are random and the intercept is selected to attain the desired proportion of missing values. This is the main setting across the experiments. 

    \item \textbf{Quantile MNAR:} 
    The missing mechanism is related to the missing variables. A subset of the remaining $70\%$ variables which will have missing variables is randomly selected. Then, missing values are generated on the $q$-quantiles at random. Since missingness depends on quantile information, it depends on masked values, hence this is a MNAR mechanism. This setting is part of our ablation studies. 
\end{itemize}

\section{Additional Experiments}\label{sup:ablation}

\begin{figure*}[h!]
    \centering
    \includegraphics[width=\linewidth]{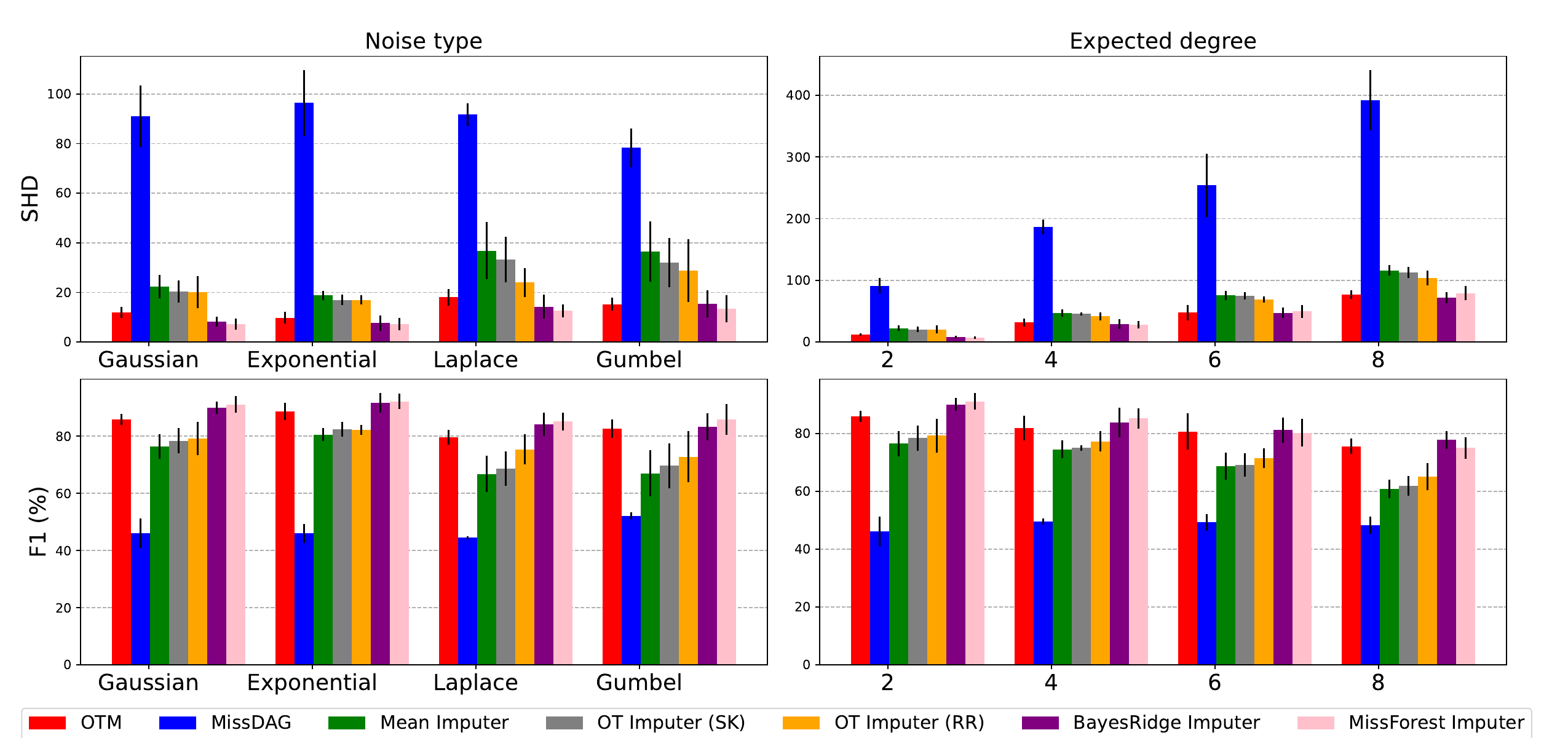}
    
    \caption{Results on different noise types and graph degrees in nonlinear ANMs (MCAR) at 10\% missing rate. \textsf{SK} refers to batch Sinkhorn imputation and \textsf{RR} refers to round-robin Sinkhorn imputation.  SHD $\downarrow$ and F1 $\uparrow$.}
    \label{fig:ablation}
\end{figure*}

\begin{figure*}
    \centering
    \includegraphics[width=\linewidth]{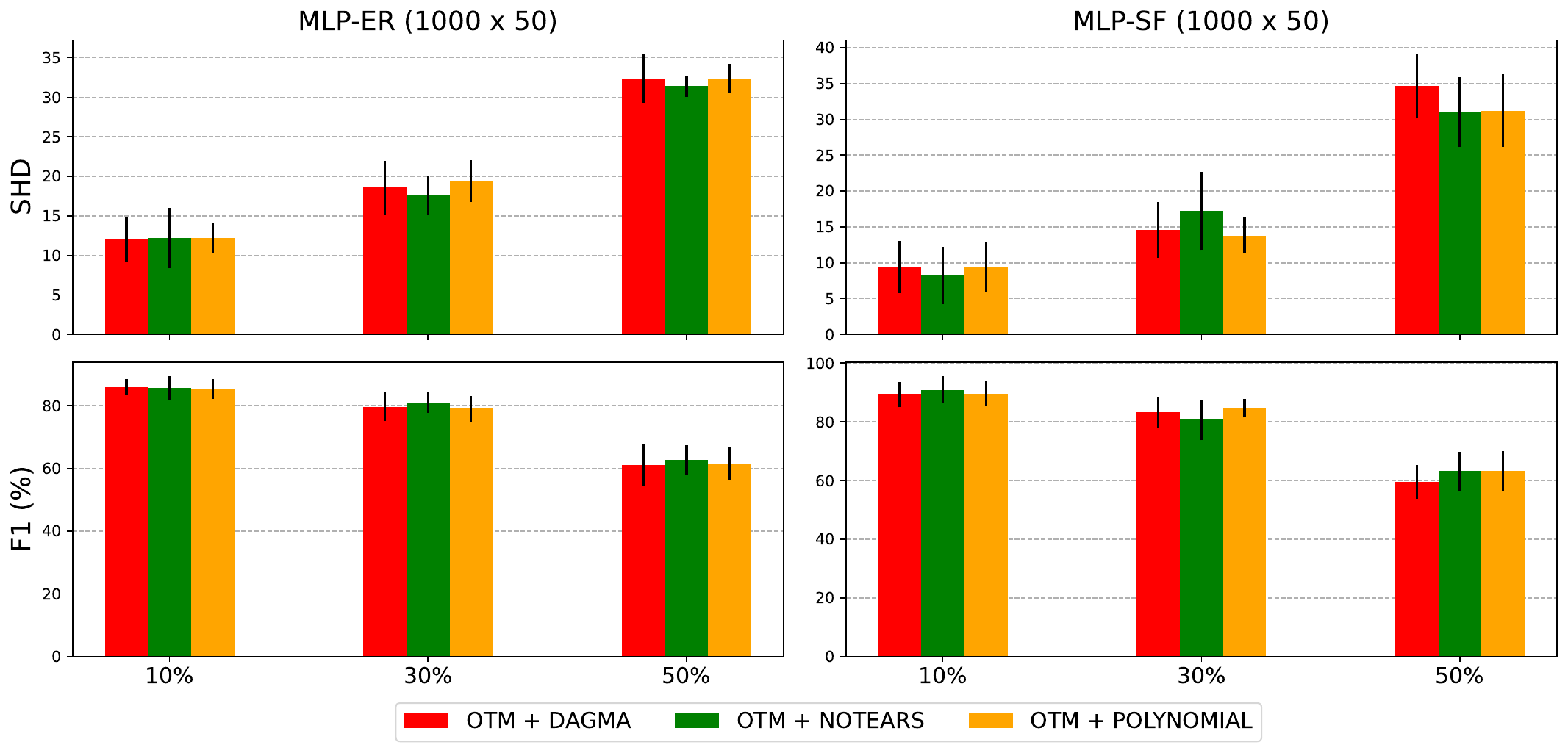}
    \caption{Results of OTM with different DAG characterizations in nonlinear ANMs (MCAR) at 10\% missing rate. SHD $\downarrow$ and F1 $\uparrow$.}
    \label{fig:reg}
\end{figure*}

In this section, we present additional results following the same setup described in Appendix \ref{sup:config}. Figures \ref{fig:sim-mar} to \ref{fig:real-mnar} illustrates the performance of methods on MAR and MNAR cases. Figure \ref{fig:ablation} and \ref{fig:reg} further investigate the methods on different configurations of noise types, graph degrees and acyclicity regularizers in MCAR cases at $10\%$ missing rate. The effectiveness of OTM remains stable across settings.   

\begin{figure*}[h!]
    \centering
    \includegraphics[width=\linewidth]{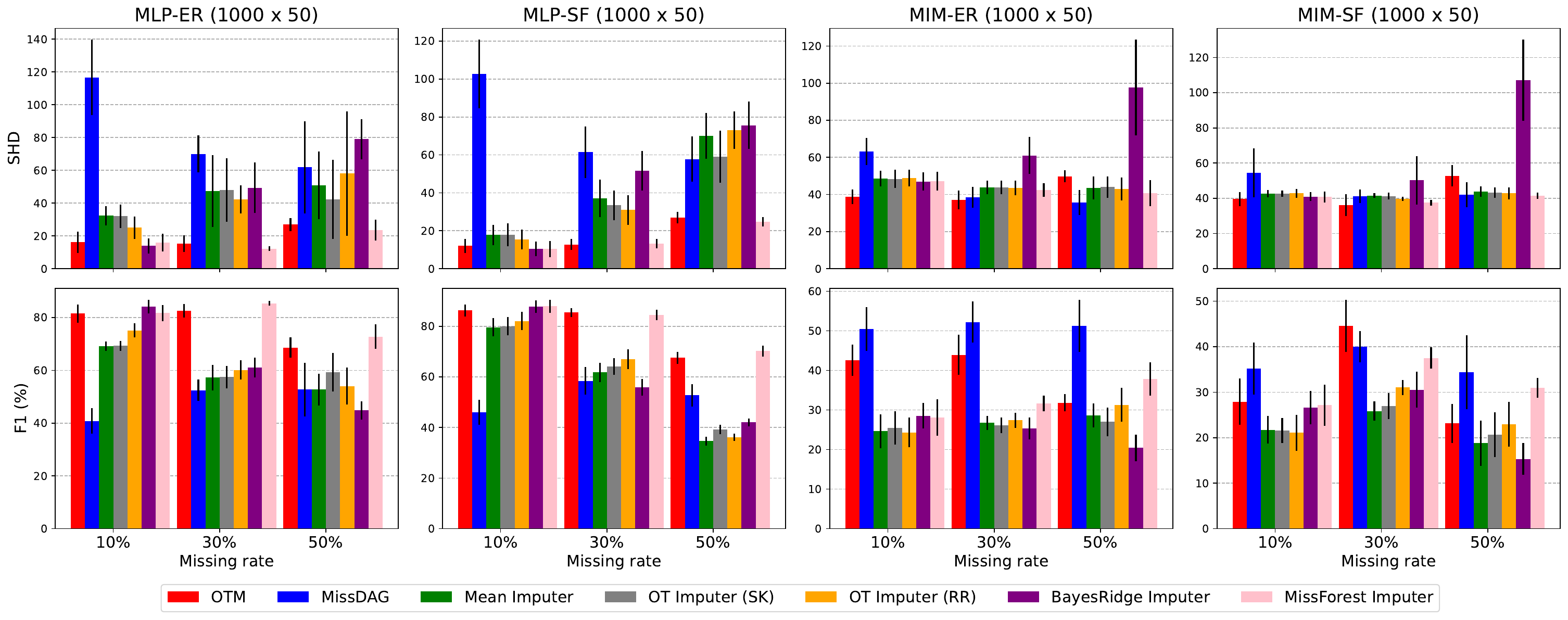}
    
    \caption{Nonlinear ANMs (MAR). \textsf{SK} refers to batch Sinkhorn imputation and \textsf{RR} refers to round-robin Sinkhorn imputation. \\ SHD $\downarrow$ and F1 $\uparrow$.}
    \label{fig:sim-mar}
\end{figure*}

\begin{figure*}[h!]
    \centering
    \includegraphics[width=\linewidth]{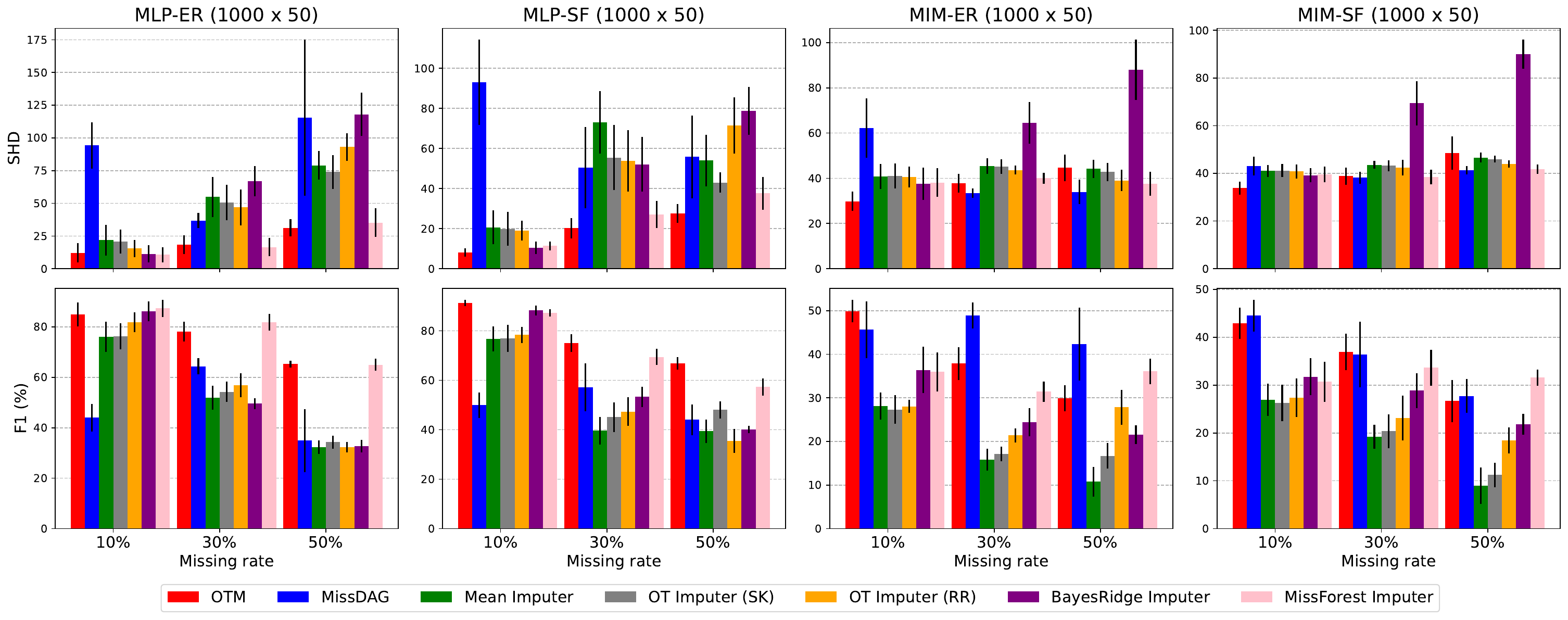}
    
    \caption{Nonlinear ANMs (Logistic MNAR). \textsf{SK} refers to batch Sinkhorn imputation and \textsf{RR} refers to round-robin Sinkhorn imputation. \\ SHD $\downarrow$ and F1 $\uparrow$.}
    \label{fig:sim-mnar}
\end{figure*}

\begin{figure*}[h!]
    \centering
    \includegraphics[width=\linewidth]{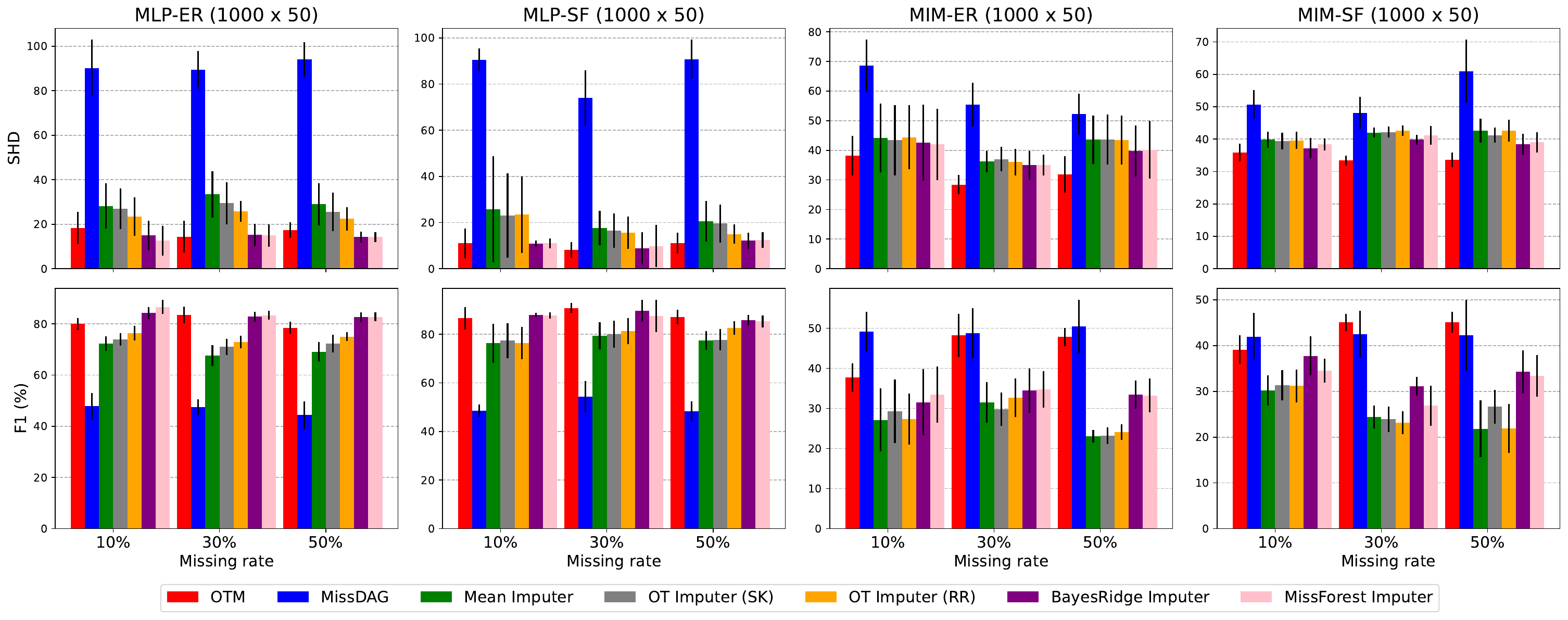}
    
    \caption{ Nonlinear ANMs (Quantile MNAR). \textsf{SK} refers to batch Sinkhorn imputation and \textsf{RR} refers to round-robin Sinkhorn imputation. \\ SHD $\downarrow$ and F1 $\uparrow$.}
    \label{fig:sim-mnar}
\end{figure*}

\begin{figure*}[h!]
    \centering
    \includegraphics[width=\linewidth]{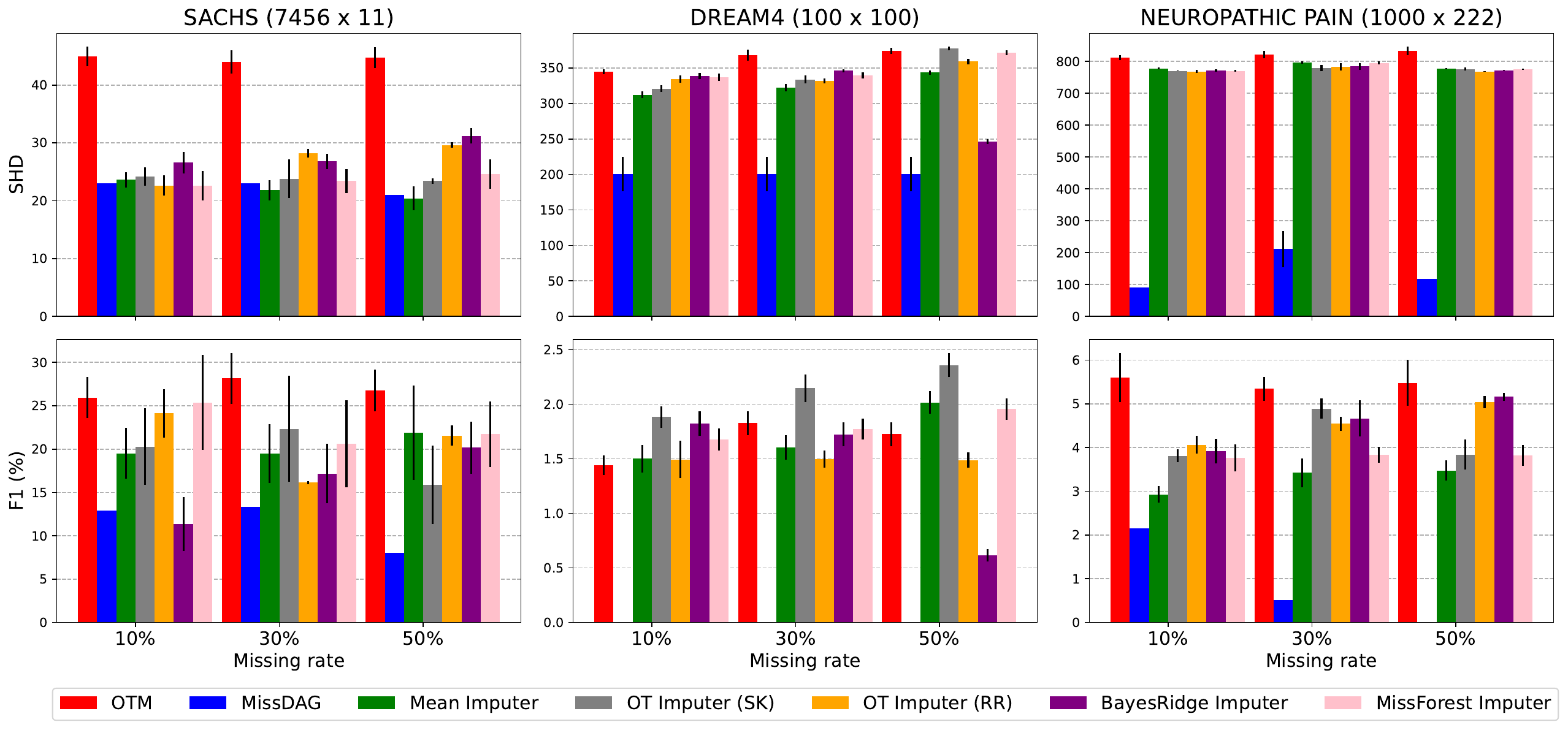}
    
    \caption{Real-world datasets (MAR). \textsf{SK} refers to batch Sinkhorn imputation and \textsf{RR} refers to round-robin Sinkhorn imputation. \\ SHD $\downarrow$ and F1 $\uparrow$.}
    \label{fig:real-mar}
\end{figure*}

\begin{figure*}[h!]
    \centering
    \includegraphics[width=\linewidth]{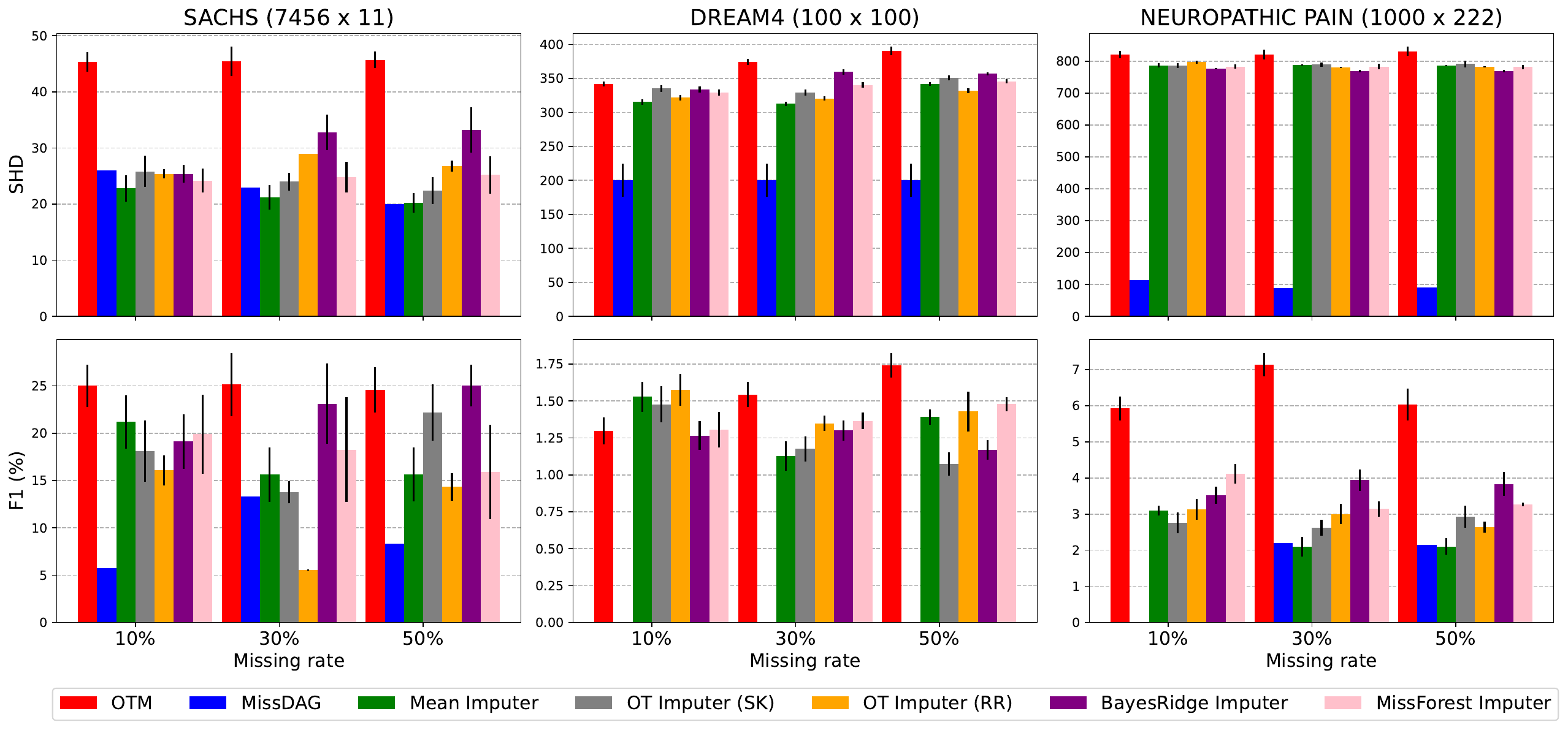}
    
    \caption{Real-world datasets (Logistic MNAR). \textsf{SK} refers to batch Sinkhorn imputation and \textsf{RR} refers to round-robin Sinkhorn imputation. SHD $\downarrow$ and F1 $\uparrow$.}
    \label{fig:real-mnar}
\end{figure*}

\paragraph{Linear ANMs.}\label{sup:linear}

\begin{figure*}[h!]
    \centering
    \includegraphics[width=\linewidth]{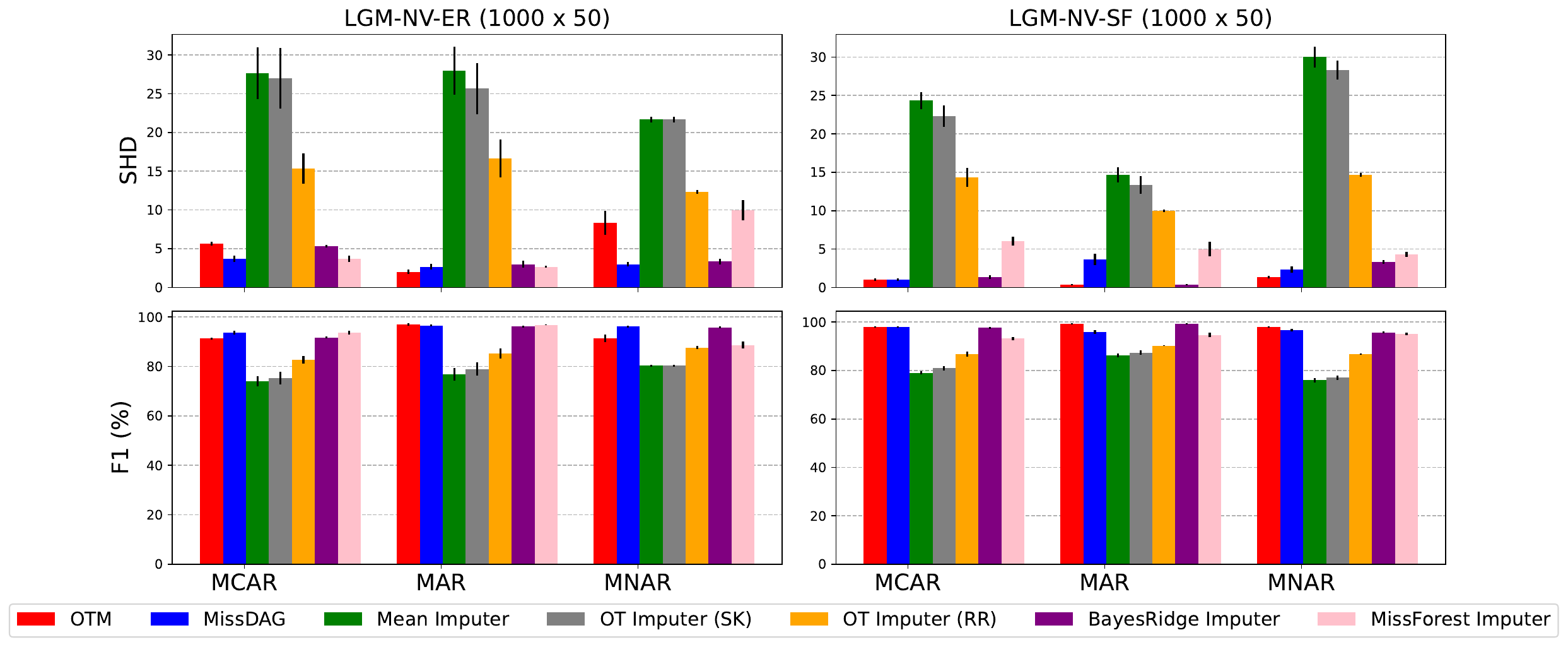}
    
    \caption{Linear ANMs at 10\% missing rate. \textsf{SK} refers to batch Sinkhorn imputation and \textsf{RR} refers to round-robin Sinkhorn imputation. \\ SHD $\downarrow$ and F1 $\uparrow$.}
    \label{fig:linear}
\end{figure*}

This section further reports the performance of methods in linear cases. We specifically study linear Gaussian models with non-equal variances (LGM-NV). For linear models, we run OTM using NOTEARS \citep{zheng2018dags} as the underlying causal discovery method to show that OTM can be flexibly integrated into any existing DAG learning methods. We use the default setting of NOTEARS (with L-BFGS-B optimizer) and model the imputation network with a simple deterministic linear layer. Figures \ref{fig:linear} demonstrates the competitive performance of OTM against MissDAG. This is achieved without explicitly doing posterior inference. 

\end{document}